\documentclass[sigconf,nonacm]{acmart}
\AtBeginDocument{%
  }

\setcopyright{acmlicensed}
\copyrightyear{2025}
\acmYear{2025}
\acmDOI{XXXXXXX.XXXXXXX}
\acmConference[Conference acronym 'XX]{Make sure to enter the correct
  conference title from your rights confirmation email}{June 03--05,
  2018}{Woodstock, NY}
\acmISBN{978-1-4503-XXXX-X/2018/06}

\usepackage[utf8]{inputenc} 
\usepackage[T1]{fontenc}    
\usepackage{hyperref}       
\usepackage{xurl}
\usepackage{booktabs}       
\usepackage{amsfonts}       
\usepackage{nicefrac}       
\usepackage{microtype}      
\usepackage{xcolor}         
\usepackage{wrapfig}
\usepackage{enumitem}
\usepackage{tikz}

\usepackage{subcaption}
\usepackage{caption}

\usepackage{cleveref}

\definecolor{mydarkblue}{rgb}{0,0.08,0.45}
\hypersetup{ %
pdftitle={},
pdfsubject={},
pdfkeywords={},
pdfborder=0 0 0,
pdfpagemode=UseNone,
colorlinks=true,
linkcolor=mydarkblue,
citecolor=mydarkblue,
filecolor=mydarkblue,
urlcolor=mydarkblue,
}

\usepackage{algorithm}
\usepackage{algpseudocode}

\usepackage{booktabs}      
\usepackage{multirow}      
\usepackage{array}         
\usepackage{tabularx}
\usepackage{makecell}  

\usepackage{tikz}
\usetikzlibrary{matrix}             

\usepackage{amsmath}
\usepackage{mathtools}
\usepackage{amsthm}
\usepackage{thmtools}
\usepackage{thm-restate}

\theoremstyle{plain}
\newtheorem{theorem}{Theorem}[section]

\theoremstyle{definition}
\theoremstyle{definition}

\newtheoremstyle{bolditalicstyle}  
  {3pt}                            
  {3pt}                            
  {\itshape}                       
  {}                               
  {\bfseries}                      
  {.}                              
  {.5em}                           
  {}                               

\theoremstyle{bolditalicstyle}

\DeclareMathOperator*{\argmax}{argmax}
\DeclareMathOperator*{\argmin}{argmin}

\newcommand{\method}[1]{\texttt{#1}}
\newcommand{\dataset}[1]{\textsf{\textit{#1}}}
\newcommand{\afabench}{\textbf{AFABench}}

\begin{document}

\title{AFABench: A Generic Framework for Benchmarking Active Feature Acquisition}


\author{Valter Schütz}
\authornote{These authors contributed equally to this work.} 
\affiliation{%
  \institution{Chalmers University of Technology \& University of Gothenburg}
  \country{Sweden}}
\email{valter.schutz@chalmers.se}

\author{Han Wu}
\authornotemark[1]                                    
\affiliation{%
  \institution{Chalmers University of Technology}
  \country{Sweden}}
\email{hanwu@student.chalmers.se}

\author{Reza Rezvan}
\affiliation{%
  \institution{Chalmers University of Technology}
  \country{Sweden}}
\email{rezvan@student.chalmers.se}

\author{Linus Aronsson}
\affiliation{%
  \institution{Chalmers University of Technology \& University of Gothenburg}
  \country{Sweden}}
\email{linaro@chalmers.se}

\author{Morteza Haghir Chehreghani}
\affiliation{%
  \institution{Chalmers University of Technology \& University of Gothenburg}
  \country{Sweden}}
\email{morteza.chehreghani@chalmers.se}

\renewcommand{\shortauthors}{Schütz et al.}

\begin{abstract}
In many real-world scenarios, acquiring all features of a data instance can be expensive or impractical due to monetary cost, latency, or privacy concerns. \textit{Active Feature Acquisition} (AFA) addresses this challenge by dynamically selecting a subset of informative features for each data instance, trading predictive performance against acquisition cost. While numerous methods have been proposed for AFA, ranging from myopic information-theoretic strategies to non-myopic reinforcement learning approaches, fair and systematic evaluation of these methods has been hindered by a lack of standardized benchmarks. In this paper, we introduce \afabench, the first benchmark framework for AFA. Our benchmark includes a diverse set of synthetic and real-world datasets, supports a wide range of acquisition policies, and provides a modular design that enables easy integration of new methods and tasks. We implement and evaluate representative algorithms from all major categories, including static, myopic, and reinforcement learning-based approaches. To test the lookahead capabilities of AFA policies, we introduce a novel synthetic dataset, \dataset{CUBE-NM}, designed to expose the limitations of myopic selection. Our results highlight key trade-offs between different AFA strategies and provide actionable insights for future research. The benchmark code is available at: \url{https://github.com/Linusaronsson/AFA-Benchmark}.
\end{abstract}

\maketitle

\section{Introduction} \label{section:introduction}

In many real-world applications, acquiring feature values of data instances can be costly. In healthcare, different medical tests can have monetary cost, latency, or privacy concerns \citep{10.5555/1622826.1622838}. In recommender systems, querying user preferences can impose a cognitive burden or intrude on privacy \citep{Jeckmans2013}. Similar constraints arise in sensor systems and robotics, where measurements require power, time, or motion \citep{10.5555/1641503.1641516, 9899480}. In such cases, acquiring the full set of features for every data instance may be infeasible or inefficient. \emph{Active Feature Acquisition} (AFA) addresses this challenge by learning to dynamically select, for each data instance, a small subset of informative features to acquire, optimizing predictive performance under a constrained acquisition budget.

Unlike static feature selection, which selects the same set of features for every example, AFA enables instance-wise decisions: the features selected for one instance can differ from those selected for another. This flexibility makes AFA more realistic and efficient in domains where the informativeness or relevance of features varies between instances. Moreover, since acquisition is sequential, AFA policies can condition future acquisition decisions on previously observed feature values, mirroring expert decision-making strategies such as in clinical diagnosis, where test results guide further examination.

Despite its relevance and importance, AFA has until recently received less attention than adjacent topics such as active learning \citep{settles.tr09}. In active learning, all features are typically available, and the goal is to selectively acquire costly labels. In contrast, AFA assumes that both labels and features are available during training, but only partial features can be acquired at test time. AFA is also different from static feature selection methods, which fix a global subset of features regardless of the test instance. For comprehensive reviews of traditional feature selection methods, we refer the reader to \citep{DBLP:journals/jmlr/GuyonE03, DBLP:journals/csur/LiCWMTTL17, DBLP:journals/ijon/CaiLWY18}.

A recent survey by \citet{aronsson2025surveyactivefeatureacquisition} categorizes AFA methods into several paradigms. At a high level, the literature can be divided into (i) \emph{myopic methods}, which select features by maximizing a one-step utility without explicit long-term planning, and (ii) \emph{non-myopic methods}, which optimize long-term utility and are often formulated within a reinforcement learning (RL) framework. By learning policies that account for future acquisitions, non-myopic methods can yield lookahead strategies that outperform myopic selection in certain regimes. \Cref{section:methods} details the AFA methods considered in this work. A central question in this benchmark is whether the additional computational cost of non-myopic planning is justified relative to simpler myopic heuristics.



Although this body of work has grown considerably in recent years, progress is hindered by the lack of a standardized and unified evaluation framework. Most existing methods are evaluated in isolation using inconsistent datasets, model architectures, or acquisition costs. To our knowledge, \textbf{no prior work has provided a dedicated benchmark for the AFA setting}. This makes it difficult to perform fair comparisons or gain generalizable insights.

This paper introduces the first benchmark for AFA. We present a unified and extensible framework for comparing AFA methods under controlled, fair, and reproducible conditions. The benchmark supports a diverse collection of synthetic and real-world datasets and includes representative algorithms across the main methodological categories: myopic approaches (both generative and discriminative), non-myopic approaches (including model-free RL, model-based RL, and non-RL variants), as well as static baselines. Crucially, the framework is \textbf{modular and easy to extend}, enabling researchers to incorporate new methods, acquisition strategies, and datasets with minimal engineering effort. This makes our framework a valuable tool for both researchers developing new AFA techniques and practitioners seeking to apply AFA in domain-specific tasks.

Our main contributions are:
\begin{itemize}[leftmargin=1.2em]
    \item We introduce \afabench, the \emph{first benchmark} for Active Feature Acquisition, allowing standardized and fair comparisons across a wide range of methods and settings.
    \item Our framework is \textbf{modular and extensible}, making it easy to add new AFA methods, datasets, and evaluation protocols.
    \item We implement and evaluate representative methods from all major paradigms, including myopic methods (generative and discriminative), non-myopic methods based on RL (model-free and model-based), non-myopic methods without explicit RL, and static feature selection.
    \item We propose a novel synthetic dataset, \dataset{Cube-NM}, that highlights the limitations of myopic strategies and provides a testbed for evaluating non-myopic acquisition policies (which we justify theoretically, see \Cref{thm:cube-nm}).
\end{itemize}

To promote reproducibility and future development, our benchmark is open source and available at: \url{https://github.com/Linusaronsson/AFA-Benchmark}. It contains step-by-step instructions on how to extend the benchmark with new models and new datasets.

\section{Problem Formulation} \label{sec:problem}

In this section, we introduce the AFA problem and the relevant notation used throughout the paper.

\subsection{Notation}

Let \(\mathbf{x} = \{\mathbf{x}_1, \ldots, \mathbf{x}_d\} \in \mathcal{X}\) denote a data instance with \(d\) features, and let \(\mathbf{y} \in \mathcal{Y}\) be the corresponding response variable in a supervised learning setting. We assume that \((\mathbf{x}, \mathbf{y})\) is distributed according to a joint data distribution \(p(\mathbf{x}, \mathbf{y})\). Throughout, we use bold symbols (e.g., \(\mathbf{x}\), \(\mathbf{y}\)) to represent random variables, and their non-bold counterparts (e.g., \(x\), \(y\)) to denote specific realizations sampled from the distribution, i.e., \(x, y \sim p(\mathbf{x}, \mathbf{y})\). In practice, each \( \mathbf{x}_i \) may correspond to a single feature (as is typical for tabular data), but it may also represent a group of features, for example a patch of pixels in an image. See Section~\ref{section:framework} for details. For any subset \(S \subseteq [d] \triangleq \{1, \ldots, d\}\), we write \(\mathbf{x}_S = \{\mathbf{x}_i \mid i \in S\} \in \mathcal{X}_S\) to denote the corresponding subset of features. For an instance $x$, acquiring feature $i \in [d]$ reveals $x_i$ and incurs the cost $c_i \in \mathbb{R}^+$. In addition, we denote $c(S) \triangleq \sum_{i \in S} c_i$ for $S \subseteq [d]$.




\subsection{Active Feature Acquisition} \label{section:afaformulation}

As discussed in Section \ref{section:introduction}, the AFA problem appears in many variants. In this section, we describe a common formulation adopted for our benchmark \citep{aronsson2025surveyactivefeatureacquisition}. In AFA, one learns (possibly jointly) two functions: (i) a predictor \(f(S, x_S)\) that produces an output given any observed subset \(S \subseteq [d]\) (typically implemented via feature masking, see, e.g., \citep{aronsson2025surveyactivefeatureacquisition}), and (ii) an acquisition policy \(\pi(S,x_S) \in \mathcal{A}_B(S)\), where
\begin{equation}\label{eq:actionspace}
    \mathcal{A}_B(S) \triangleq \{\, a \in U : c(S) + c_a \le B \,\} \cup \{\texttt{STOP}\},
\end{equation}
is the set of allowed actions after acquiring $S$. Here, \(U = [d] \setminus S\) denotes the set of unobserved features, and \(B\) is a per-instance budget on cumulative feature cost. In practice, the policy often outputs a distribution over actions. If \(\pi\) ouputs a feature acquisition action \(a \in U\), we incur cost \(c_a\) and observe \(x_a\). We then update $S$ and $U$ as follows: \(S \leftarrow S \cup \{a\}\) and \(U \leftarrow U \setminus \{a\}\). If instead \(\pi\) selects \texttt{STOP}, we output \(\hat{y} = f(S, x_S)\) and terminate. Let \(\pi[x] \subseteq [d]\) denote the final set of acquired feature indices when the policy stops on instance \(x\). Given this, the optimization objective for AFA is shown below.
\begin{equation}
\begin{aligned}
\label{eq:afaopt}
&\min_{f,\pi} 
\mathbb{E}_{\mathbf{x},\mathbf{y}} 
\mathbb{E}_{\pi} \left[
\ell \big(f(\mathbf{x}_{\pi[\mathbf{x}]}),\mathbf{y}\big)
 + \alpha c(\pi[\mathbf{x}])\right] \\
&\quad\text{s.t.}\quad
c(\pi[x])\le B\;\;\text{for all }x\in\mathcal{X}.
\end{aligned}
\end{equation}
where \( \alpha \ge 0 \) controls the cost--accuracy trade-off and \( \ell \) is a loss function. The term \( \alpha\, c(\pi[\mathbf{x}]) \) allows the acquisition cost to vary across instances, enabling the policy to spend more resources on instances that are harder to predict. 

\textbf{Hard vs. soft budget}. From the above, \( \alpha \) induces a \emph{soft} budget and is therefore referred to as a \emph{soft-budget parameter}. By contrast, \( B \) imposes a \emph{hard} budget. A soft budget is beneficial because it adapts the cost spent per instance. However, selecting an appropriate value of \( \alpha \) can be unintuitive in many applications (as discussed by, e.g., \citep{DBLP:journals/ml/JanischPL20}). A hard budget \( B \) is more natural when resources are strictly limited per instance and is often easier to interpret. Its drawback is that some instances may require acquiring features whose total cost exceeds \( B \) to achieve accurate predictions.

\textbf{A note on terminology}. Throughout the paper, the term \emph{AFA method} refers to the predictor $f$ \emph{and} the policy $\pi$ induced by the method. Some methods learn $f$ and $\pi$ jointly, for instance by training $f$ on the states $(S, x_S)$ encountered under $\pi$ during learning. Other methods instead pretrain $f$ on randomly sampled states $(S, x_S)$ and then use this predictor when training $\pi$ (optionally continuing to update $f$ using states encountered under $\pi$). This distinction is discussed further in \Cref{section:external_vs_internal_predictor}.


\subsection{Scope of the Benchmark} \label{section:scope}

Most prior work on AFA considers two special cases of~\eqref{eq:afaopt}: (i) the \emph{hard-budget} setting, where \( \alpha = 0 \) (only a hard budget), and (ii) the \emph{soft-budget} setting, where \( B = c([d]) \) (all features are, in principle, acquirable for every instance, so only the soft-budget term matters). Following prior work, we also focus on these special cases in our benchmark.

We focus the scope of our benchmark on methods developed for the common \emph{offline} setting, where the goal is to deliver accurate but cost-effective predictions at \emph{test time}. In this setting, one assumes access to a fully observed training dataset \(\mathcal{D} = \{(x_1,y_1),\dots,(x_N,y_N)\}\) of \(N\) i.i.d.\ samples  
\((x_i,y_i) \sim p(\mathbf{x},\mathbf{y})\).  
A predictor \(f\) and an acquisition policy \(\pi\) are learned on \(\mathcal{D}\). Then, for each test instance, the policy \(\pi\) is used to acquire features, and the predictor \(f\) produces a prediction based solely on the acquired subset. There are several different methods for learning $f$ and $\pi$ based on $\mathcal{D}$. 

Finally, we narrow the scope to classification, as this is the main focus of previous work \citep{aronsson2025surveyactivefeatureacquisition}.

\subsection{MDP Formulation of AFA} \label{section:mdpformulation}

The AFA problem can be formulated as a \emph{partially observable Markov decision process} (POMDP) \citep{aronsson2025surveyactivefeatureacquisition}. Equivalently, it can be expressed as a fully observable MDP by defining the state space as \( \mathcal{S} = \{(S, x_S) : S \in 2^{[d]},\, x_S \in \mathcal{X}_S\} \), where a state comprises the set \( S \) of acquired features together with their observed values \( x_S \) \citep{aronsson2025surveyactivefeatureacquisition}. The action space and transition dynamics are exactly as described in Section~\ref{section:afaformulation}. In the standard MDP formulation of AFA, the reward is \( r((S,x_S), a) = -\alpha c_a \) for \( a \in \mathcal{A}_B(S) \setminus \{\texttt{STOP}\} \) and \( r((S,x_S), \texttt{STOP}) = -\ell(f(S,x_S), y) \). Given this, it is straightforward to show that maximizing expected return in this MDP is equivalent to \eqref{eq:afaopt}. This connection was first established by \citep{dulac2011datum}, where model-free reinforcement learning (RL) was used to learn non-myopic policies for solving \eqref{eq:afaopt}.

Some RL methods in this benchmark adopt alternative reward functions, for example by shaping intermediate rewards to reflect the information gained about the label \( y \), in order to mitigate sparse-reward issues \citep{kachuee2018opportunistic, zannone2019odin}. Such shaping can, however, bias learning toward more myopic acquisition strategies. The (PO)MDP underlying AFA is known to be highly intractable (expecially for large $d$), motivating the development of myopic heuristics. We refer to \citep{aronsson2025surveyactivefeatureacquisition} for details. The benchmark in this paper investigates whether the additional computational cost of non-myopic planning is justified relative to simpler myopic heuristics.

\section{A General Framework for Implementing AFA Methods} \label{section:framework}

We now describe how we implement the AFA problem introduced in \Cref{section:afaformulation} in our benchmark. The framework described here applies primarily to evaluation, since AFA methods may be trained in arbitrary ways, but consistent evaluation is essential for fair benchmarking.

\begin{figure}[t]
	\centering
	\includegraphics[width=0.5\textwidth]{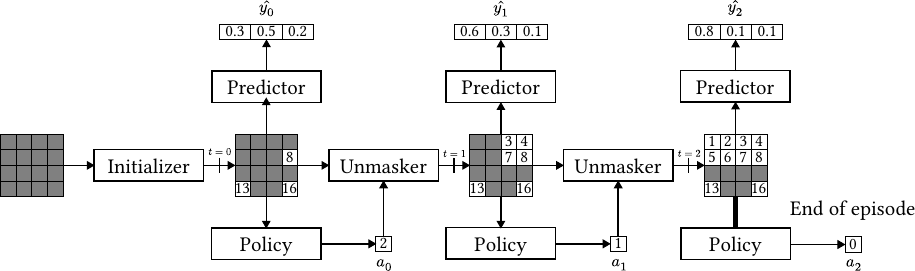}
	\caption{An AFA episode where the instance $x \in \mathcal{X}$ is an image.}
	\label{fig:afa_episode}
\end{figure}

To benchmark AFA methods in a standardized manner, we must make explicit assumptions about their inputs and outputs. During evaluation, the components of the AFA problem in our benchmark interact episodically for each instance $x \in \mathcal{X}$ in the dataset. An example episode is illustrated in \Cref{fig:afa_episode}, where the instance $x \in \mathcal{X}$ is a $4 \times 4$ feature matrix representing an image.

An \texttt{Initializer} determines which features are observed at the start of an episode, using either deterministic or stochastic rules. The resulting masked instance is then passed in parallel to a \texttt{Predictor} and a \texttt{Policy}. The \texttt{Predictor} produces an intermediate prediction $\hat{y}_t$, which is stored for subsequent performance evaluation but does not otherwise affect the episode dynamics. In parallel, the \texttt{Policy} outputs an action $a_t$, interpreted either as a stop decision or as a request to reveal additional feature(s) (see \eqref{eq:actionspace}). 

Our framework does not require a one-to-one correspondence between actions and the features to be revealed. Instead, this mapping is handled by a separate component, the \texttt{Unmasker}, which translates actions into the next set of features to reveal. In \Cref{fig:afa_episode}, at time step $t=0$, the \texttt{Policy} outputs $a_0 = 2$, which, given the image setting and the current \texttt{Unmasker}, corresponds to the upper-right $2 \times 2$ patch. After the \texttt{Unmasker} reveals the selected features, the time index is incremented and the episode continues until the policy selects the stop action or the feature budget is exceeded.

Acquiring features in groups and doing so sequentially (one group per step), rather than selecting individual features, is a common strategy for mitigating the computational intractability of the AFA-MDP in high-dimensional settings \citep{gadgil2024estimating, aronsson2025surveyactivefeatureacquisition}. Consequently, when a user adds a new dataset with a different acquisition scheme, they only need to implement an \texttt{Unmasker} that specifies the corresponding masking and unmasking rules.

\subsection{Cost Accumulation}
\begin{figure}[t]
	\centering
	\includegraphics[width=0.3\textwidth]{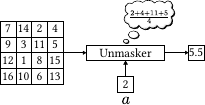}
	\caption{An example of how an \texttt{Unmasker} could calculate the cost of an action, given feature costs from a dataset.}
	\label{fig:afa_cost_calculation}
\end{figure}

An episode may terminate when the cumulative cost $c(\pi[x])$ exceeds the budget $B$ (see \eqref{eq:afaopt}). Each action selected by the \texttt{Policy} incurs a cost that depends on both the dataset and the \texttt{Unmasker}, as illustrated in \Cref{fig:afa_cost_calculation}. We assume that each dataset specifies a fixed cost per feature. Since a single action may reveal multiple features, the \texttt{Unmasker} is responsible for determining the set of features revealed by an action and summing their costs to obtain the action cost. Costs are accumulated over time steps. If the \texttt{Policy} outputs an action that would cause the budget to be \textbf{exceeded}, the action is overridden and treated as a stop action.

\subsection{External vs. internal predictors}
\label{section:external_vs_internal_predictor}

Although \Cref{fig:afa_episode} depicts the \texttt{Predictor} and \texttt{Policy} as separate components, some approaches train them jointly \citep{aronsson2025surveyactivefeatureacquisition}. While joint training can improve performance, it conflates gains due to improved action selection with gains due to improved prediction. To disentangle these effects, we compute two predictions during evaluation: one produced by an external classifier and one produced by the method's internal predictor (when available). We focus on the results based on the external classifier, since it is available to all methods and enables a more fair comparison.

\section{AFABench: An AFA Benchmark} \label{section:benchmark}
\afabench{} is a benchmark of AFA methods that leverages the implementation framework described in \Cref{section:framework}. It is written in Python and PyTorch \cite{pytorch} for training and comparing AFA methods under different conditions. It supports multiple datasets, initializers, unmaskers, predictors, and policies. Furthermore, the whole pipeline is executable with a single command, using Snakemake \cite{snakemake}.

As the main focus of this work is to benchmark feature acquisition policies, neither the \texttt{Initializer} nor the \texttt{Unmasker} is investigated in-depth. More specifically, only a \textit{cold start} \texttt{Initializer} is used, which always sets the initial feature mask to all zeros. A \textit{direct} \texttt{Unmasker} is used for almost every dataset, which maps each action to a single feature. For \dataset{CUBE-NM}, which includes explicit context features, we additionally use a \textit{context} \texttt{Unmasker} with a dedicated action that acquires all context features at once. \dataset{ImageNette} also uses a different, patch-based image \texttt{Unmasker}, like the one in \Cref{fig:afa_episode}. Similarly, since benchmarking different types of classifiers is not the goal, a simple MLP classifier is used for all datasets except ImageNette, where a Vision Transformer (ViT) is used \cite{dosovitskiy2021an}.

\subsection{Pipeline Stages}

As mentioned, the whole pipeline can be executed with a single command, lowering the barrier to adoption. In order to achieve this, some assumptions have to be made about how the policies. We assume that the construction of every policy can be split into a pretraining and training stage (both optional), followed by a mandatory evaluation stage, which uses the same script for all methods and implements the framework described in \Cref{section:framework}. For example, consider a method that relies on a pretrained external classifier. The pipeline is then as follows. First, pretrain the external classifier. Next, train the policy $\pi$ using this classifier. Finally, evaluate the resulting policy (and the corresponding predictor) on test instances. In addition, some methods (e.g., AACO \citep{pmlr-v235-valancius24a}) do not require an explicit training stage, since all computations are performed at inference time (i.e, non-parametric, see \Cref{appendix:methods} for details). The pipeline stages are illustrated in \Cref{fig:pipeline}. 

\begin{figure}
    \centering
    \usetikzlibrary{positioning}
\begin{tikzpicture}[>=stealth, node distance=15mm, thick, scale=0.4, transform shape]

\newcommand{\TikzFontComp}{\fontsize{20}{30}\selectfont}

\tikzstyle{stage}=[rectangle, draw=black, minimum width=40mm, minimum height=20mm, font=\TikzFontComp\bfseries, line width=0.8mm, align=center]
\tikzstyle{optional}=[rectangle, draw=black, dashed, minimum width=40mm, minimum height=20mm, font=\TikzFontComp\bfseries, line width=0.8mm, align=center]
\tikzstyle{arrow}=[->, black, line width=0.8mm]
\tikzstyle{component}=[font=\TikzFontComp\bfseries, align=center]
\tikzstyle{graycomponent}=[font=\TikzFontComp\bfseries, align=center, text=gray]

\node[optional] (pretraining) {Pretraining};
\node[optional, right=30mm of pretraining] (training) {Training};
\node[stage, right=30mm of training] (evaluation) {Evaluation};

    \node[component, above=5mm of training] (afapredictor2) {Predictor};
    \node[component, below=5mm of training] (afapolicy2) {Policy};
    \node[component, above=5mm of evaluation] (afapredictor3) {Predictor};
    \node[component, below=5mm of evaluation] (afapolicy3) {Policy};
    
\draw[arrow] (pretraining) to (training);
\draw[arrow] (training) to (evaluation);
\end{tikzpicture}
    \caption{The pipeline consists of three stages; optional stages are indicated by dotted lines.}
    \label{fig:pipeline}
\end{figure}

Furthermore, the hard budget $B$ and the soft-budget parameter $\alpha$ are specified at both training and evaluation time (and may differ between the two). This is necessary because some methods use only the soft-budget parameter during training (for example, RL-based methods), whereas others use it only at evaluation time (for example, myopic methods based on~\eqref{eq:cmi}). More generally, one can combine an arbitrary choice of soft and/or hard budgets at training with an arbitrary choice of soft and/or hard budgets at evaluation, to accommodate any kind of AFA method.

All hard budgets and soft-budget parameters are listed in configuration files, making it easy to run additional experiments without requiring knowledge about the code base.

\subsection{Included AFA Methods} \label{section:methods}

As described in the recent survey by \citet{aronsson2025surveyactivefeatureacquisition}, AFA methods can be broadly categorized as follows:
(i) \emph{myopic} heuristics, which perform no explicit planning beyond a one-step lookahead and are often further classified as \emph{discriminative} or \emph{generative},
(ii) \emph{non-myopic} methods based on reinforcement learning (RL), which learn acquisition policies by exploiting the MDP structure underlying AFA (see \Cref{section:mdpformulation}) and can be divided into \emph{model-free} and \emph{model-based} approaches, and
(iii) \emph{non-myopic} methods that do not rely on explicit RL.

All myopic methods included in this benchmark acquire the next feature $a \in \mathcal{A}_B(S) \setminus \{\texttt{STOP}\}$ by maximizing the expected information gain about the label (relative to its cost):
\begin{equation} \label{eq:cmi}
    \pi_{\text{CMI}}(S,x_S) = \argmax_{a \in \mathcal{A}_B(S) \setminus \{\texttt{STOP}\}} I(\mathbf{y};\mathbf{x}_a \mid x_S)/c_a.
\end{equation}
Here, $I(\mathbf{y};\mathbf{x}_a \mid x_S) = H(\mathbf{y} \mid x_S) - \mathbb{E}_{\mathbf{x}_a \mid x_S}\!\left[H(\mathbf{y} \mid x_S, \mathbf{x}_a)\right]$ denotes the conditional mutual information (CMI) between $\mathbf{y}$ and $\mathbf{x}_a$. The myopic methods differ only in how this CMI term is estimated (see \Cref{appendix:methods}). Their myopic nature is evident in that they ignore interactions between $\mathbf{x}_a$ and the remaining unobserved features in $U$, that is, they perform no planning beyond a one-step lookahead.

The benchmark includes at least one representative state-of-the-art method from each category to provide a broad comparison of AFA strategies. Table~\ref{tab:feature-taxonomy} summarizes the methods included in the benchmark. Among the myopic methods, EDDI-GM \citep{ma2019eddi} estimates the CMI using a generative model, whereas GDFS-DM \citep{covert2023learning} and DIME-DM \citep{gadgil2024estimating} estimate CMI using discriminative predictors. The model-free RL-based methods include: JAFA-MFRL \citep{NEURIPS2018_e5841df2}, OL-MFRL \citep{kachuee2018opportunistic}, and ODIN-MFRL \citep{zannone2019odin}. ODIN-MBRL \citep{zannone2019odin} uses model-based RL by utilizing an explicit model for planning. AACO \citep{pmlr-v235-valancius24a} represents a non-myopic approach that does not explicitly rely on RL, instead using oracle-based lookahead to guide acquisitions. Finally, PT-S \citep{DBLP:journals/ml/Breiman01} and CAE-S \citep{DBLP:conf/icml/BalinAZ19} are static feature selection baselines based on global feature importance. Detailed descriptions of all implemented AFA methods are provided in \Cref{appendix:methods}.



\subsection{Included Datasets}
The benchmark covers both synthetic and real-world tasks across several
domains: \dataset{CUBE}, a widely used synthetic dataset for AFA, and
\dataset{CUBE-NM}, a new synthetic dataset introduced in this work
(\Cref{section:syntheticdata}); \dataset{MNIST}, handwritten digit
recognition \citep{DBLP:journals/pieee/LeCunBBH98}, and
\dataset{FashionMNIST}, grayscale clothing-item classification
\citep{xiao2017/online}, both treated as tabular problems by viewing
each pixel as a feature; \dataset{Diabetes}, a diabetes diagnosis task
derived from NHANES data \citep{NHANES2018} and commonly used in AFA
\citep{kachuee2018opportunistic, covert2023learning};
\dataset{PhysioNet}, ICU mortality prediction based on EHR data from
the PhysioNet Challenge 2012 \citep{Goldberger2000PhysioNet};
\dataset{MiniBooNE}, a particle identification dataset from the
MiniBooNE experiment at Fermilab \citep{miniboone_particle_identification_199};
\dataset{ACTG175}, a clinical trial dataset from the UCI repository
\citep{ACTG175}; \dataset{CKD} (chronic kidney disease) and
\dataset{BankMarketing}, both standard UCI tabular datasets
\citep{CKD,bank_marketing}; and \dataset{Imagenette}, a 10-class subset
of ImageNet used here to evaluate patch-based image acquisition
\citep{Imagenette}. Additional details and preprocessing are
provided in \Cref{appendix:datasets}. See Table \ref{tab:datasets} for a
summary of the datasets.

\subsection{CUBE-NM: A Novel Synthetic Dataset for AFA} \label{section:syntheticdata}
\begin{figure*}[t]
  \centering
  \begin{subfigure}[t]{0.4\textwidth}
      \centering
      \includegraphics[width=\linewidth]{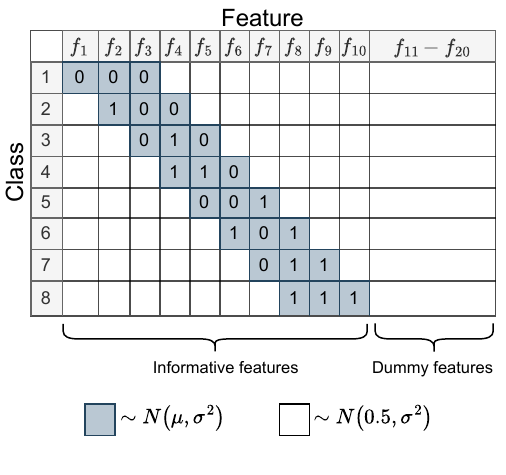}
      \caption{\textbf{CUBE}.}
      \label{fig:plot-a}
  \end{subfigure}
  \hfill
  \begin{subfigure}[t]{0.55\textwidth}
    \centering
    \includegraphics[width=\linewidth]{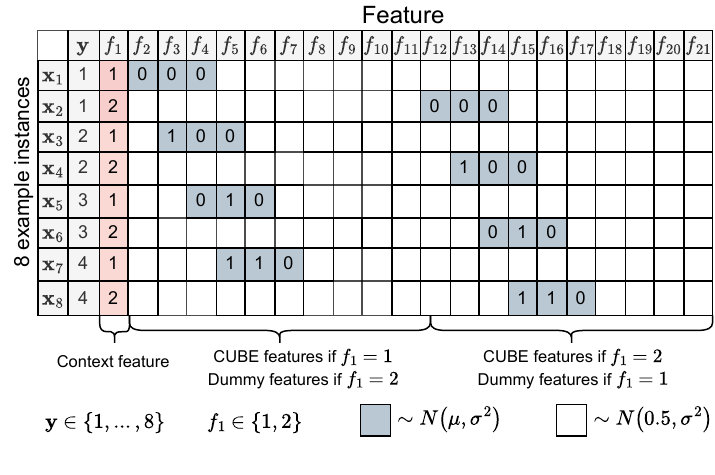}
    \caption{\textbf{Cube-NM} for $n_c = 2$.}
    \label{fig:plot-b}
  \end{subfigure}
  \caption{Visualization of (a) \textbf{CUBE}, and (b) our proposed synthetic dataset \dataset{Cube-NM}.}
  \label{fig:side-by-side-plots}
\end{figure*}

In this section we introduce \dataset{CUBE-NM}, a dataset for benchmarking non-myopic methods for AFA. We first recall the widely used \dataset{CUBE} dataset \citep{ruckstiest2013minimizing, NEURIPS2018_e5841df2, zannone2019odin}, illustrated in Figure~\ref{fig:plot-a}. \dataset{CUBE} consists of 20-dimensional real-valued vectors from 8 classes. For each class, three informative features are drawn from a Gaussian distribution \( \mathcal{N}(\mu,\sigma^2) \) with class-specific mean \( \mu \), while all remaining features are sampled as noise from \( \mathcal{N}(0.5,\sigma^2) \). The locations of the informative features (and their mean) depend on the class (colored cells in \Cref{fig:plot-a}, where the numbers indicate the corresponding means). This structure makes \dataset{CUBE} well suited for evaluating instance-specific (dynamic) acquisition, but it offers limited separation between myopic and non-myopic methods (shown formally in Theorem \ref{thm:cube-nm}).


\dataset{CUBE-NM} extends \dataset{CUBE} by adding a context feature that selects which block of features is informative, while the context value itself is not predictive under myopic criteria. Concretely, each instance has a single integer-valued context feature \(f_1\in\{1,\dots,n_c\}\). For each context \(j\in[n_c]\), we associate a block of 10 real-valued features \(B_j=\{\,f_{1+10(j-1)+r} : r=1,\dots,10\,\}\). Conditional on \(f_1=j\), the features in \(B_j\) follow the same informative structure as \dataset{CUBE} (with \( \mathcal{N}(\mu,\sigma^2) \) for informative coordinates and \( \mathcal{N}(0.5,\sigma^2) \) for noise coordinates). The total dimensionality is \(d=1+10n_c\). Figure~\ref{fig:plot-b} shows example instances for the case \(n_c=2\). An optimal policy should therefore acquire \(f_1\) to identify the active block, and then query within that block.

This design creates a setting where myopic methods struggle: the key first step is to acquire \(f_1\), but \(f_1\) provides no immediate label information, since \(I(\mathbf{y};\mathbf{x}_{f_1})=0\) (see \eqref{eq:cmi}). As long as there exist other candidate features with \(I(\mathbf{y};\mathbf{x}_a)>0\), myopic methods will keep selecting within blocks without first identifying the correct one. Non-myopic methods can instead pay a small upfront cost to observe \(f_1\), after which they can focus queries in the informative block. Theorem \ref{thm:cube-nm} formalizes the benefit of a non-myopic approach on this dataset.
\begin{theorem}[Informal]\label{thm:cube-nm}
Consider the CUBE-NM dataset with $n_c$ contexts in the noiseless regime $\sigma=0$ (such that $100\%$ prediction accuracy is possible). Assume uniform unit acquisition costs, i.e., $c_i=1$ for all $i\in[d]$. Then the myopic CMI policy in~\eqref{eq:cmi} requires, in expectation across instances, $13(2n_c+1)/16$ feature acquisitions to achieve $100\%$ accuracy. In contrast, there exists an optimal (non-myopic) policy that achieves $100\%$ accuracy after acquiring only $\mathbf{1}\{n_c\ge 2\}+9/4$ features in expectation.
\end{theorem}
A formal statement and proof of \Cref{thm:cube-nm} is included in \Cref{appendix:proofcube}. For \(n_c=1\) and \(\sigma=0\), \dataset{CUBE-NM} reduces to \dataset{CUBE}, since \(f_1\) is constant and can be ignored by any method (and the noise features \(f_{11}-f_{20}\) in \dataset{CUBE} have a negligible effect when \(\sigma\) is small). In this case, \Cref{thm:cube-nm} shows that the myopic method is nearly optimal: the myopic and optimal policies require $2.25$ and $2.4375$ feature acquisition in expectation, respectively. As $n_c$ increases, the gap grows linearly. For example, for $n_c=3$ we obtain expected costs $3.25$ and $5.6875$ for the myopic and optimal policy, respectively. \Cref{fig:cube_nm_three_panels} illustrates the difference for more values of $n_c$. Thus, \(n_c\) tunes the benefit of non-myopic acquisition. A useful property is that increasing \(n_c\) also increases the dimensionality \(d=10n_c+1\), making it harder to learn a policy that reliably exploits the larger gap. Finally, while \Cref{thm:cube-nm} assumes $\sigma=0$, the general principle holds for reasonable noise levels $\sigma > 0$.

\subsection{Ensuring a Fair Benchmark} \label{section:fair-benchmark}

In this section, we summarize the key assumptions needed for a fair and objective benchmark. In \Cref{appendix:fair-benchmark}, we include more details about each point below. Notably, nearly all prior AFA studies violate at least one of the assumptions below, most often the first.

\textbf{Hard vs.\ soft budget}. We evaluate soft-budget and hard-budget settings separately, since existing AFA methods often target only one and mixing them can confound comparisons. Where possible, methods are adapted to match each objective. In \Cref{appendix:methods}, we describe how each method is adapted to both setting. Most existing AFA work does not make this distinction, which can confound results by comparing methods that optimize two different objectives.

\textbf{Common components}. All methods are implemented in a unified framework with shared components kept consistent across methods. For example, \method{EDDI-GM} and the \method{ODIN} variants use the same pretrained partial VAE (see \Cref{appendix:methods}). In addition, in our experiments using a shared external predictor (see next point), all methods use the same pretrained predictor.

\textbf{Internal vs.\ external predictor}. Because some methods jointly learn $f$ with the policy $\pi$ while others assume a fixed $f$, we report two settings: (i) evaluation with a shared external predictor for all methods, and (ii) evaluation with the method’s internal predictor when available.

\textbf{Consistent RL framework}. RL-based methods share the same overall training loop, differing only in the RL algorithm and reward definition. Implementation details are provided in \Cref{appendix:methods}.

\textbf{Excluded methods}. We exclude approaches that are tied to specific predictor classes or rely on strong distributional assumptions, and instead focus on model-agnostic methods applicable across domains. See the survey in \citep{aronsson2025surveyactivefeatureacquisition} for details on other methods.

\textbf{Randomness}. Results are averaged over multiple data splits and runs, and we report standard deviations (shown as error bars) to reflect variability.

\textbf{Adherence to original architecture}. We primarily use the hyperparameters and architectures from original papers or repositories, since exhaustive tuning across all settings is impractical. When necessary (e.g., to avoid clear underfitting, we make adjustments). See \Cref{appendix:experiments} for more hyperparameter details.



\section{Experiments}
\label{section:experiments}

As discussed in \Cref{section:benchmark}, we analyze the soft-budget and hard-budget settings separately. In \Cref{appendix:methods}, we detail how each method is adapted to each setting, where we also describe how static methods are adapted to AFA for our experiments. Here, consistent with most prior work on AFA \citep{aronsson2025surveyactivefeatureacquisition}, we focus on uniform cost, that is, $c_i = 1$ for all $i \in [d]$. In \Cref{section:cube_nm_heatmap}, we additionally report results for non-uniform cost. We report accuracy for class-balanced datasets and F1-score for class-imbalanced datasets. For the \dataset{CUBE-NM} dataset, we use $n_c = 5$ and $\sigma=0$. For \dataset{CUBE} we use $\sigma=0.3$. All results are averaged over five train/validation/test splits. 

In \Cref{section:results_soft_budget}, we study the soft-budget setting, in which policies are allowed to stop early and no hard budget is enforced at evaluation time. Instead, each method's stopping behavior is governed by its soft-budget parameter. For each method--dataset pair, we tune this parameter so that the average number of selections spans a representative range. If the soft-budget parameter cannot be tuned reliably for a given dataset, we exclude the corresponding method from that dataset. In \Cref{section:results_hard_budget}, we study the hard-budget setting, in which policies have no incentive to stop early and therefore always select up to the fixed budget $B$. During evaluation, all methods are constrained by the same hard budget and are not provided with any soft-budget parameter $\alpha$. Finally, \Cref{section:cube_nm_heatmap} visualizes the actions selected by different methods on the \dataset{CUBE-NM} dataset, introduced in \Cref{section:syntheticdata}.

We only report performance achieved at the end of each episode, in both the hard-budget and the soft-budget settings, in accordance with \eqref{eq:afaopt}. As discussed in \Cref{section:benchmark}, we evaluate each method using both (i) a shared external predictor and (ii) a method-specific internal predictor. For brevity, in the main text we only present results for the external predictor, which was trained on the first split of each dataset. We defer the internal predictor results to \Cref{appendix:moreresults}. For the same reason, we include only a subset of datasets here, and provide complete results across all datasets in \Cref{appendix:moreresults}. 

We report variance only for the final performance metric (variance on the $y$-axis). In the soft-budget setting, there is also variance in the number of selected features across instances, and it is common to visualize this variance as well (i.e., variance on the $x$-axis). To keep the plots in this section as clear as possible, we defer results that include variance in the number of selected features to \Cref{appendix:moreresults}. 

When multiple variants of a method are considered, we report only the best-performing variant in this section. A detailed comparison of method variants is presented in \Cref{appendix:method_variations}. We report the runtime of each method, split into the pretraining, training, and evaluation stages as in \Cref{fig:pipeline}, in \Cref{appendix:compute}. Finally, additional details on the experimental setup, including training details and the choice of hyperparameters, are provided in \Cref{appendix:experiments}.

\begin{figure*}[t]
    \centering
    \includegraphics[width=\textwidth]{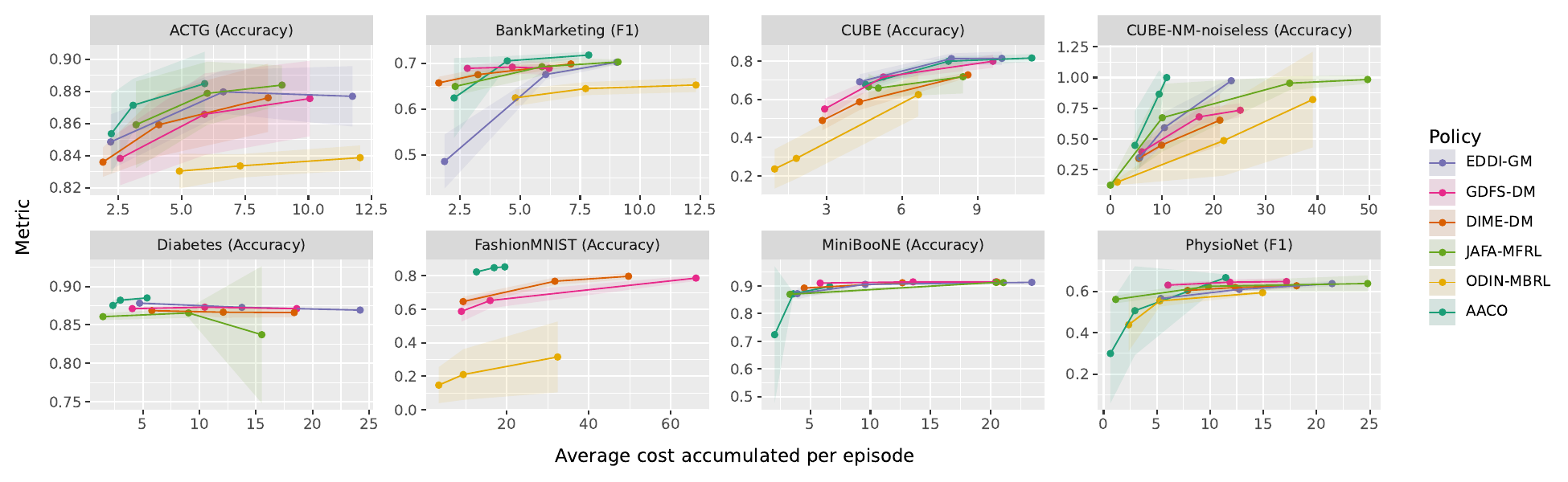}
    \caption{Results for the soft-budget setting using a shared external classifier across all methods. Not all methods are applicable in the soft-budget setting (see main text).}
    \label{fig:soft_budget_lines_main_external_set1}
\end{figure*}

\begin{figure*}[t]
    \centering
    \includegraphics[width=\textwidth]{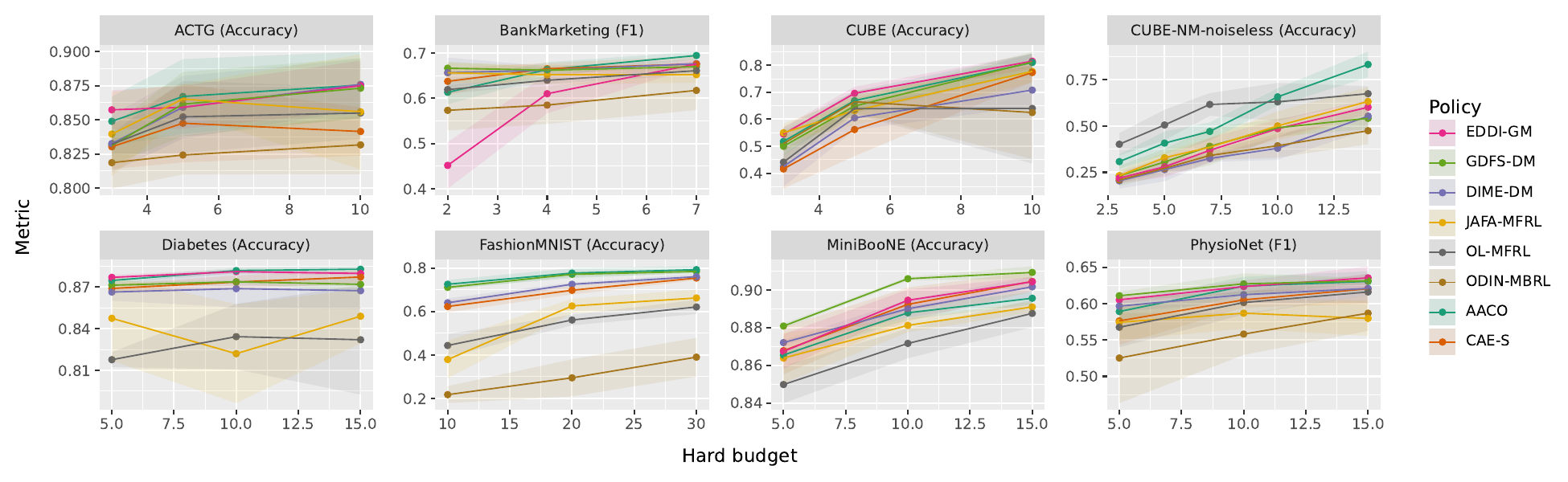}
    \caption{Results for the hard-budget setting using a shared external classifier across all methods.}
    \label{fig:hard_budget_main_external_set1}
\end{figure*}

\subsection{Soft-budget Results}
\label{section:results_soft_budget}

First, \method{OL-MF} is not included in the soft-budget setting, since adapting it to this setting is not straightforward and would require substantial deviations from its original formulation \citep{kachuee2018opportunistic}. Second, the static methods are also omitted, because static methods do not perform instance-wise selection and are therefore not applicable in the soft-budget setting. See \Cref{appendix:methods} for details. We can, however, compare with their corresponding performance in the hard-budget setting (see \Cref{fig:hard_budget_main_external_set1}). 
\Cref{fig:soft_budget_lines_main_external_set1} shows results for the soft budget setting using a shared external classifier. For each method, the soft budget parameter $\alpha$ induces a policy $\pi$ whose feature acquisition decisions may correspond to different cumulative costs across instances. We therefore map each value of $\alpha$ to the resulting cumulative cost averaged across instances and over train, validation, and test splits. This average cost is shown on the $x$-axis. 

We observe that the myopic approaches \method{GDFS-DM}, \method{DIME-DM}, and \method{EDDI-GM} are often among the best-performing methods. Interestingly, within this group, the generative approach \method{EDDI-DM} is not substantially worse than the discriminative methods \method{GDFS-DM} and \method{EDDI-GM}, which contrasts with the conclusions reported in \citep{covert2023learning, gadgil2024estimating}. This discrepancy may be explained by differences in the evaluated datasets, or by our tuning of the PVAE model used by \method{EDDI-GM} to estimate \eqref{eq:cmi} (see \Cref{appendix:methods} for details). 
However, for several datasets, including \dataset{ACTG}, \dataset{FashionMNIST}, and \dataset{CUBE-MN}, we find a clear advantage for non-myopic selection. In particular, the non-myopic method \method{AACO}, which does not rely on RL, performs well on all of these three datasets. 
In contrast, the RL-based methods, while in principle well-suited to settings with non-myopic structure, generally do not match the performance of \method{AACO}, with the exception of \method{JAFA-MFRL} on \dataset{PhysioNet}. We also find that many RL-based methods exhibit substantial variance, most notably \method{ODIN-MBRL}, suggesting unstable training dynamics. A plausible explanation is the intractability of the underlying MDP for AFA, which many of these methods attempt to solve directly (see \citep{aronsson2025surveyactivefeatureacquisition} and \Cref{section:mdpformulation}).

\begin{figure*}[thb!]
    \centering
\includegraphics[width=\textwidth]{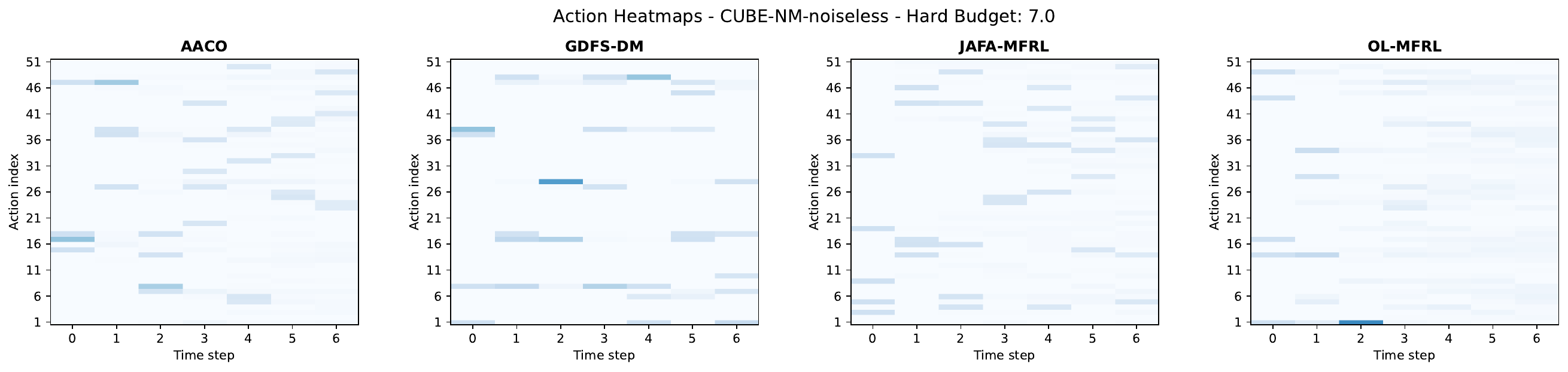}
    \caption{Heatmap of features acquired at different time steps by four methods on the CUBE-NM dataset with $\sigma=0$ and $n_c = 5$. Action index $1$ corresponds to the context feature $f_1$ (see \Cref{section:syntheticdata}).}
    \label{fig:hello}
\end{figure*}

\subsection{Hard-budget Results}
\label{section:results_hard_budget}
For the experiments in the hard-budget setting, and for readability of the results, we include only one representative static method in the main plots, namely \method{CAE-S}. Results for the second static method, \method{PT-S}, are provided in \Cref{appendix:moreresults} (it performs similarly to \method{CAE-S}). 

\Cref{fig:hard_budget_main_external_set1} shows the results for the hard-budget setting using a shared external classifier. For each method, each hard budget $B$ (shown on the $x$-axis) induces a policy $\pi$. We then use this policy to acquire features until the budget $B$ is exhausted, and report the prediction performance at the end of the episode. See \Cref{section:methods} for details on how each method is adapted to this setting.

On the \dataset{CUBE-NM} dataset, the RL-based method \method{OL-MFRL} performs best at smaller budgets. Inspecting \Cref{fig:hello} confirms that this advantage arises because it sometimes acquires the context feature (we return to this in the next subsection). We also find that the benefit of non-myopic methods over myopic methods is less pronounced here than in the soft budget setting. A likely reason is that the hard budget constrains every instance to the same total cost, which limits instance-wise adaptivity. If an optimal non-myopic strategy requires acquiring more than $B$ features for some instances to achieve high accuracy, this behavior is not feasible under a hard budget. Conversely, increasing $B$ to accommodate such instances tends to reduce the gap, since myopic methods also perform well once the budget is sufficiently large. 
Finally, static methods perform well on some datasets. For instance, \method{CAE-S} ranks among the top performers on \dataset{diabetes}, \dataset{BankMarketing}, and \dataset{PhysioNet}, suggesting that a single fixed subset of features is effective for these tasks. In contrast, on datasets such as \dataset{ACTG}, \dataset{CUBE}, and \dataset{FashionMNIST}, dynamic feature acquisition, where the selected features can differ across instances, is substantially more beneficial.

\subsection{Feature Selections on Cube-NM}
\label{section:cube_nm_heatmap}

In \Cref{fig:hello}, we visualize the features acquired by four methods on the CUBE-NM dataset with $\sigma=0$ and $n_c = 5$ at different time steps in the hard budget setting. The context feature $f_1$ corresponds to the row indexed by $1$ on the $y$-axis. See \Cref{appendix:actionheatmaps} for analogous visualizations on additional datasets.

We observe that only the RL-based method \method{OL-MFRL} learns to select the context feature consistently. However, it does not always acquire it at the first step, which would be optimal. From \Cref{fig:hard_budget_main_external_set1}, we also see that \method{OL-MFRL} achieves the best early-selection performance. In contrast, the RL-based method \method{JAFA-MFRL} never selects the context feature, and consequently performs worse. Overall, this illustrates that RL-based methods can learn non-myopic policies, but may fail to do so reliably, due to training instability. 
At the same time, the non-myopic method \method{AACO} performs strongly (and improves over the myopic baselines) despite not selecting the context feature. This indicates that \dataset{CUBE-NM} admits effective non-myopic strategies that do not rely on acquiring the context feature. Moreover, although \method{AACO} can capture non-myopic behavior, it can be shown that it will never acquire the context feature as its first action on \dataset{CUBE-NM}. Thus, \method{AACO} should be viewed as an approximation of the MDP underlying AFA, going beyond a purely myopic approximation, but not guaranteed to recover the optimal non-myopic policy. Finally, we observe that the myopic method \method{GDFS-DM} sometimes acquires the context feature. In principle, this should not occur: by construction, $I(\mathbf{y}; \mathbf{x}_{f_1}) = 0$ for the context feature $f_1$, and \method{GDFS-DM} converges exactly to the myopic conditional mutual information rule in \eqref{eq:cmi}. We therefore attribute these selections to training instability and randomness.


\section{Conclusion}
In this work, we study Active Feature Acquisition (AFA), where the goal is to balance predictive accuracy against the cost of acquiring features. To enable fair and comprehensive evaluation of AFA methods, we introduce \afabench, the first standardized benchmark framework for this task. \afabench{} includes a diverse set of synthetic and real-world datasets, supports a wide range of AFA methods, and provides a modular design to promote extensibility.

Using \afabench, we implement and evaluate representative algorithms from the major AFA paradigms, including (i) myopic methods, (ii) non-myopic methods based on RL, and (iii) non-myopic methods that do not rely on explicit RL. We additionally include static feature selection methods as simple baselines. Finally, we introduce a new synthetic dataset, \dataset{CUBE-NM}, to assess the lookahead capabilities of different methods, and we provide a theoretical characterization of the performance gap between optimal non-myopic and myopic policies on this dataset.

Our empirical analysis highlights key trade-offs among AFA strategies: (i) Although non-myopic methods are capable of learning non-myopic policies that can outperform myopic selection, this is not guaranteed, largely due to the increased complexity and potential instability of these approaches. Moreover, as our results show, many real-world datasets do not exhibit strong non-myopic structure. In such cases, myopic methods offer a strong alternative with significantly shorter training time. Consequently, future work on AFA should aim to characterize when, for a given dataset, the benefits of non-myopic planning justify its additional computational cost. (ii) Dynamic, instance-wise feature selection is often advantageous, especially for datasets with high variability across samples. However, for some datasets, static selection performs comparably well. In addition to their competitive performance, static methods are typically more efficient to train. We hope that this benchmark will serve as a basis for future research in cost-sensitive learning and more effective feature acquisition strategies.

\section*{Acknowledgments}

The work of Linus Aronsson and Morteza Haghir Chehreghani was partially supported by the Wallenberg AI, Autonomous Systems and Software Program (WASP) funded by the Knut and Alice Wallenberg Foundation. The work of Valter Schütz and Morteza Haghir Chehreghani was partially supported by the Swedish Research Council VR (grant number 2023-04809). Finally, the computations and data handling was enabled by resources provided by the National Academic Infrastructure for Supercomputing in Sweden (NAISS), partially funded by the Swedish Research Council through grant agreement no. 2022-06725.



\bibliographystyle{ACM-Reference-Format}
\bibliography{references}

\appendix

\newpage
\section{Extended Results} \label{appendix:moreresults}

\subsection{More soft-budget results}

\begin{figure*}
	\centering
	\includegraphics[width=\textwidth]{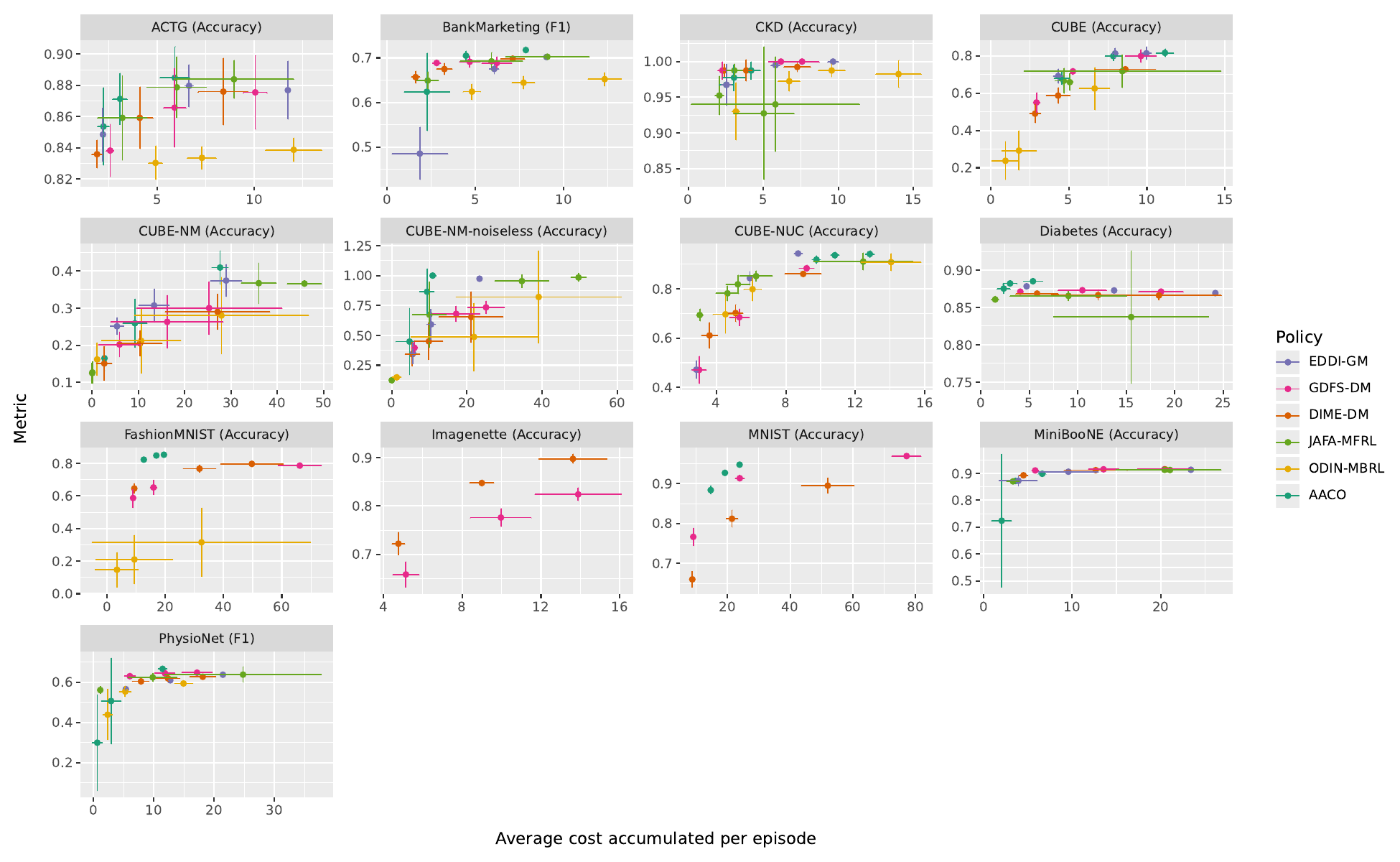}
	\caption{Soft-budget results with external classifiers}
	\label{fig:soft_budget_2d_errors_main_external_all}
\end{figure*}

\begin{figure*}
	\centering
	\includegraphics[width=\textwidth]{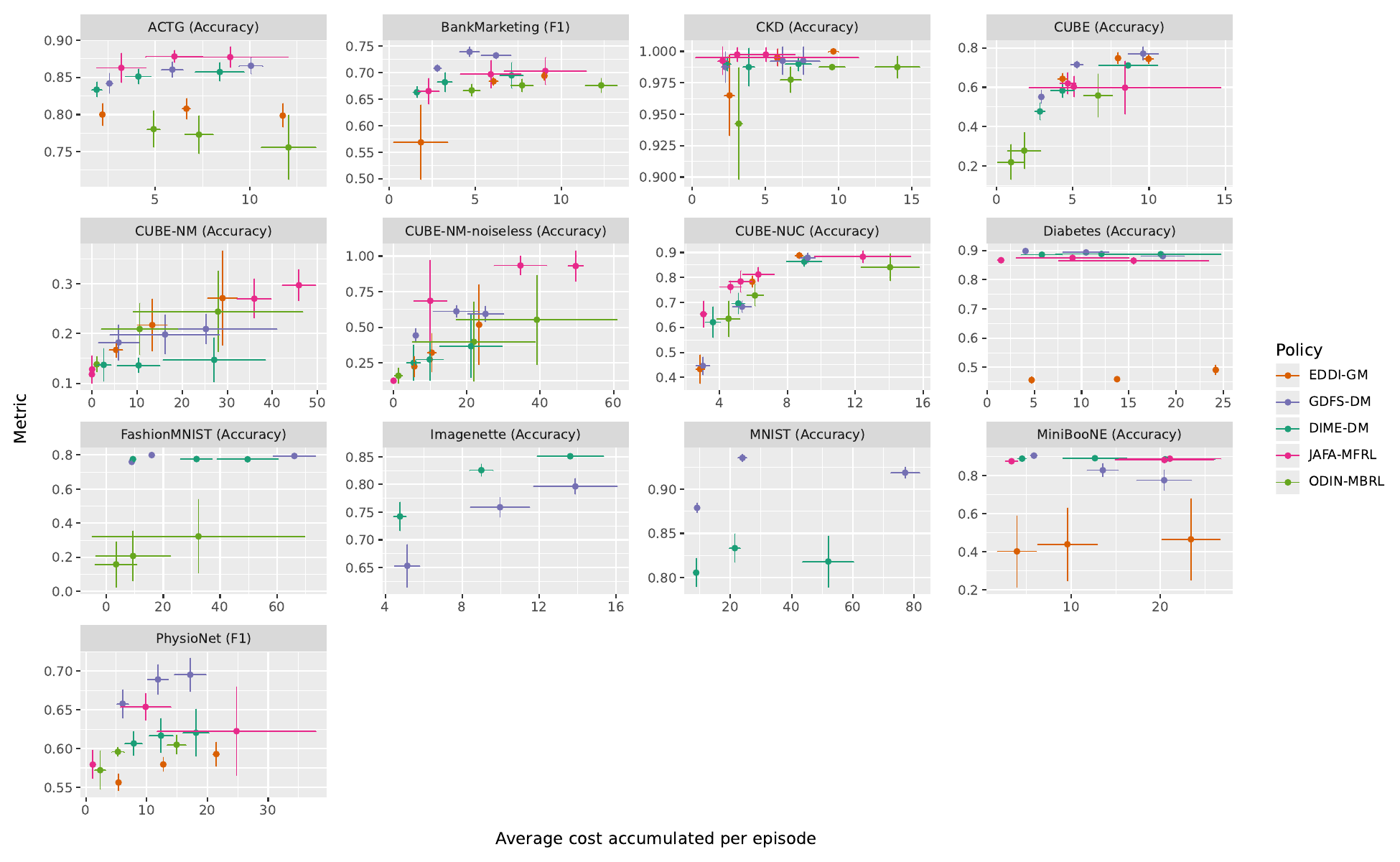}
	\caption{Soft-budget results with built-in classifiers}
	\label{fig:soft_budget_2d_errors_main_builtin_all}
\end{figure*}

\Cref{fig:soft_budget_2d_errors_main_external_all} illustrates soft-budget results using external classifiers and for the same methods as in the main text, but also includes additional datasets. \Cref{fig:soft_budget_2d_errors_main_builtin_all} shows results using built-in classifiers instead. We observe that the results are consistent with those in \Cref{fig:soft_budget_lines_main_external_set1} of the main text. We note that \method{EDDI-GM} performs poorly when using the built-in classifier on some datasets, suggesting that its performance is sensitive to the choice of classifiers.

\subsection{More hard-budget results}

\begin{figure*}
	\centering
	\includegraphics[width=\textwidth]{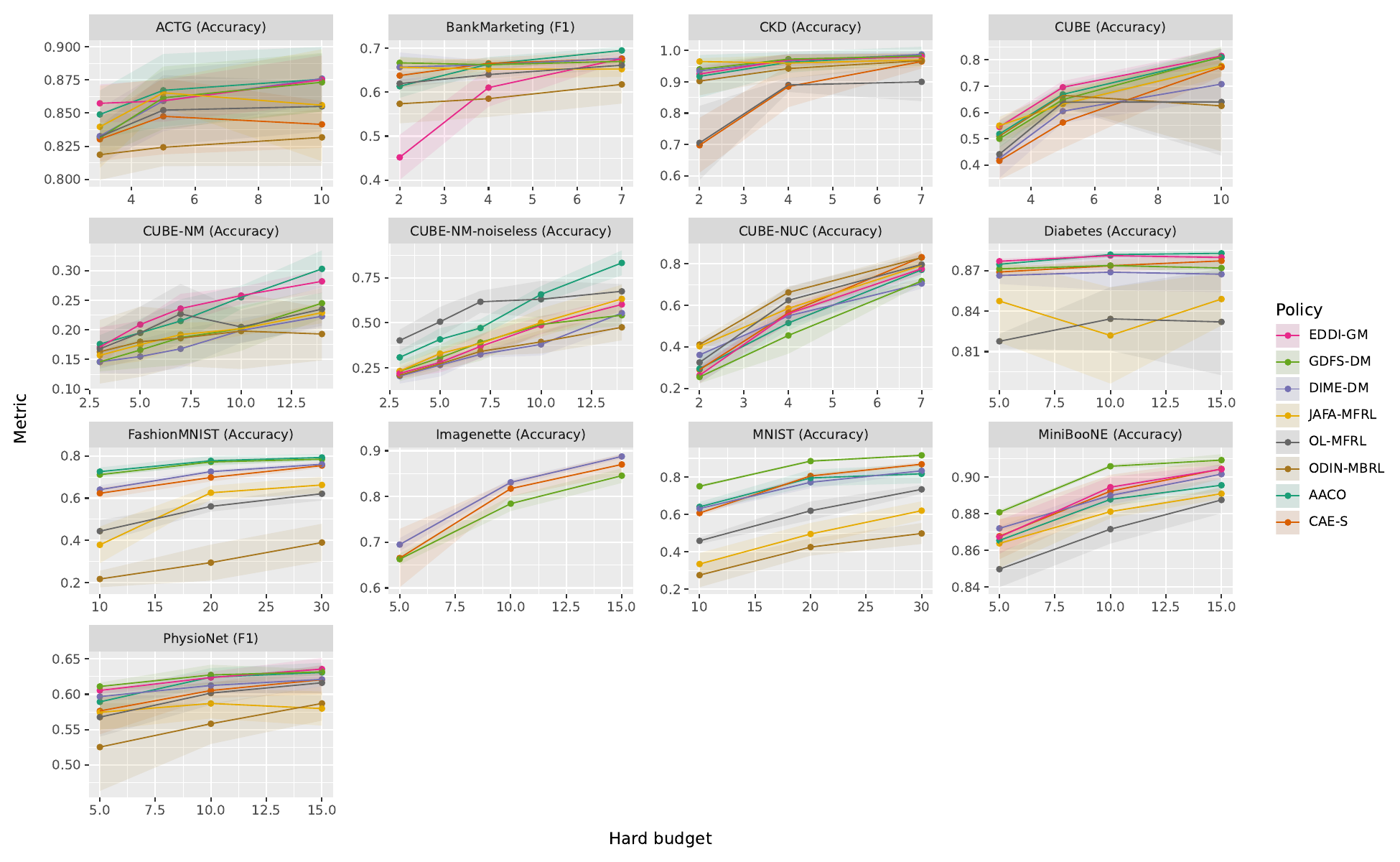}
	\caption{Hard-budget results with external classifiers.}
	\label{fig:hard_budget_main_external_all}
\end{figure*}

\begin{figure*}
	\centering
	\includegraphics[width=\textwidth]{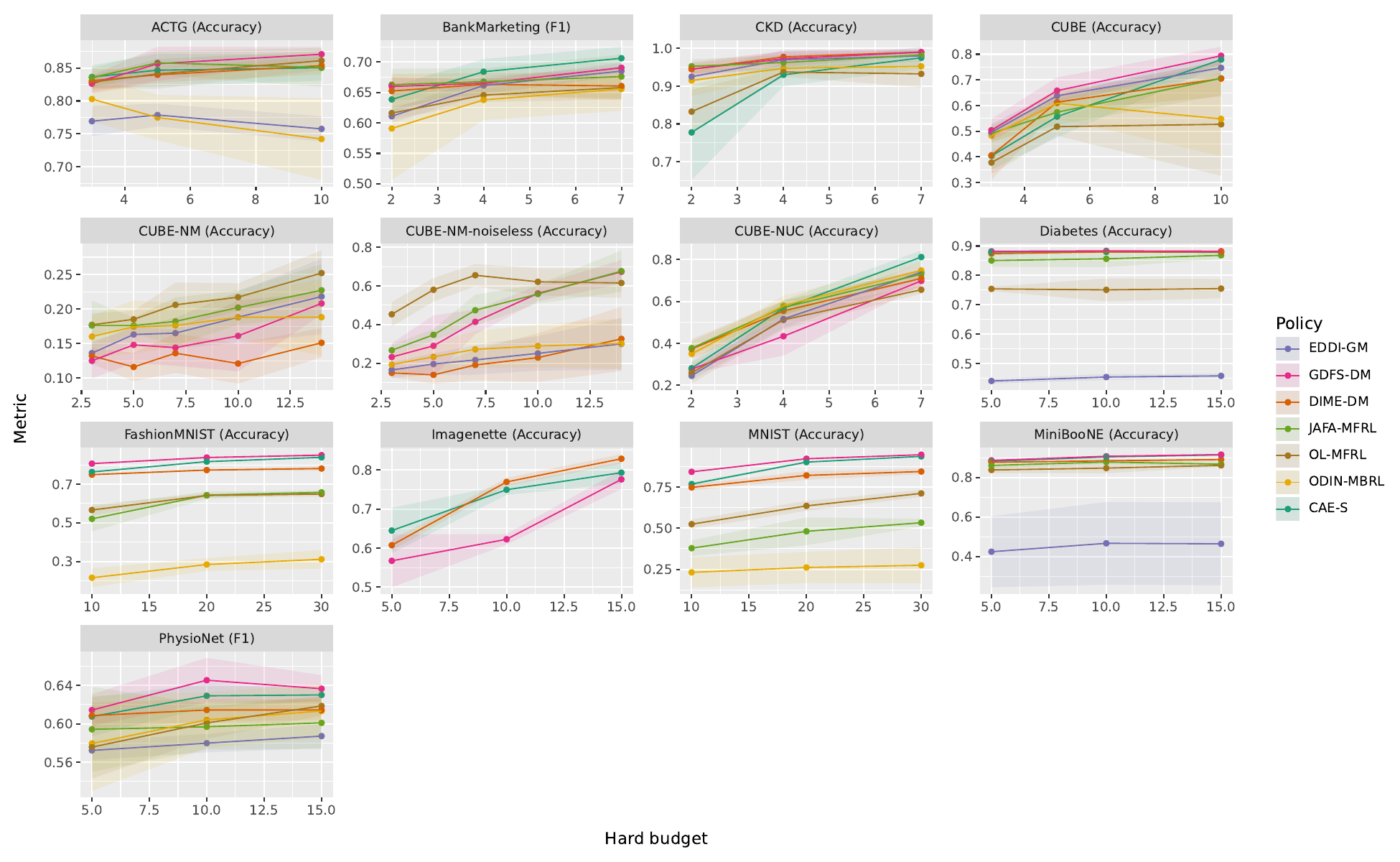}
	\caption{Hard-budget results with built-in classifiers.}
	\label{fig:hard_budget_main_builtin_all}
\end{figure*}

\Cref{fig:hard_budget_main_external_all} illustrates the hard-budget results using external classifiers and for the same methods as in the main text, but also includes additional datasets. \Cref{fig:hard_budget_main_builtin_all} shows the results using built-in classifiers instead. We observe consistent results with those in \Cref{fig:hard_budget_main_external_set1}, where myopic methods achieve competitive results on most tabular datasets, non-myopic methods achieve better results on the \dataset{CUBE-NUC} dataset, and the static method achieves competitive results on a subset of datasets. \method{EDDI-GM} shows similar sensitivity to classifier choice as in the soft-budget case.

\subsection{Action heatmaps} \label{appendix:actionheatmaps}

\subsubsection{Action heatmaps on CUBE-NM-noiseless}

\begin{figure*}
	\centering
	\includegraphics[width=\textwidth]{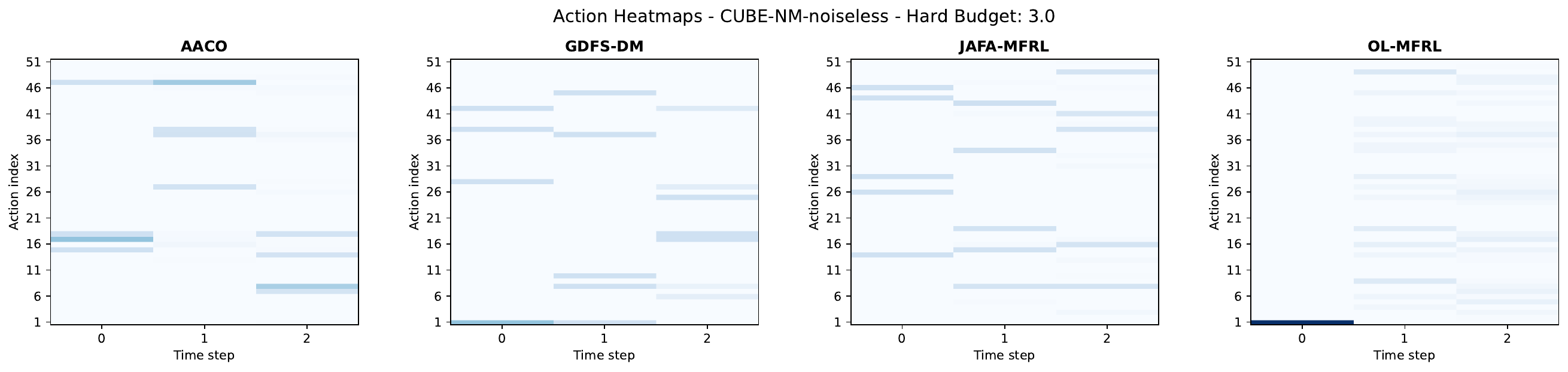}
	\caption{Heatmap of actions performed during evaluation with hard budget 3 on the \dataset{CUBE-NM-noiseless} dataset.}
	\label{fig:appendix_action_heatmap_cube_nm_noiseless_budget_3}
\end{figure*}

\begin{figure*}
	\centering
	\includegraphics[width=\textwidth]{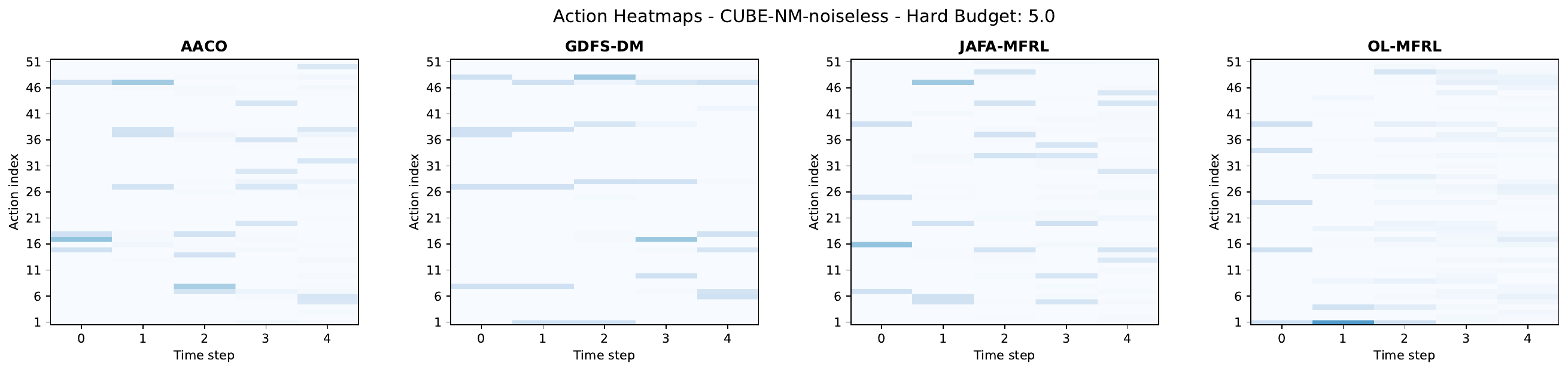}
	\caption{Heatmap of actions performed during evaluation with hard budget 5 on the \dataset{CUBE-NM-noiseless} dataset.}
	\label{fig:appendix_action_heatmap_cube_nm_noiseless_budget_5}
\end{figure*}

\begin{figure*}
	\centering
	\includegraphics[width=\textwidth]{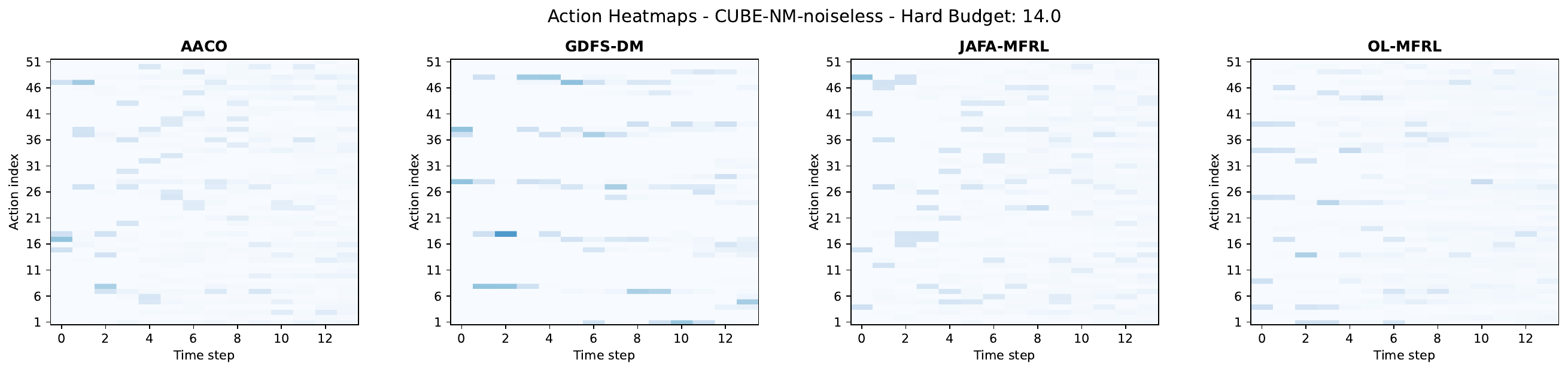}
	\caption{Heatmap of actions performed during evaluation with hard budget 14 on the \dataset{CUBE-NM-noiseless} dataset.}
	\label{fig:appendix_action_heatmap_cube_nm_noiseless_budget_14}
\end{figure*}

Extending the results from \Cref{section:cube_nm_heatmap}, \Cref{fig:appendix_action_heatmap_cube_nm_noiseless_budget_3}, \Cref{fig:appendix_action_heatmap_cube_nm_noiseless_budget_5}, and \Cref{fig:appendix_action_heatmap_cube_nm_noiseless_budget_14} contain action heatmaps of four methods on \dataset{CUBE-NM}, with three different hard budgets. We note that \method{OL-MFRL} becomes less consistent at choosing the context feature as the hard budget increases, but also that \method{GDFS-DM} picks the context feature for the smallest budget.

\subsubsection{Action heatmaps on ACTG}
\begin{figure*}
	\centering
	\includegraphics[width=\textwidth]{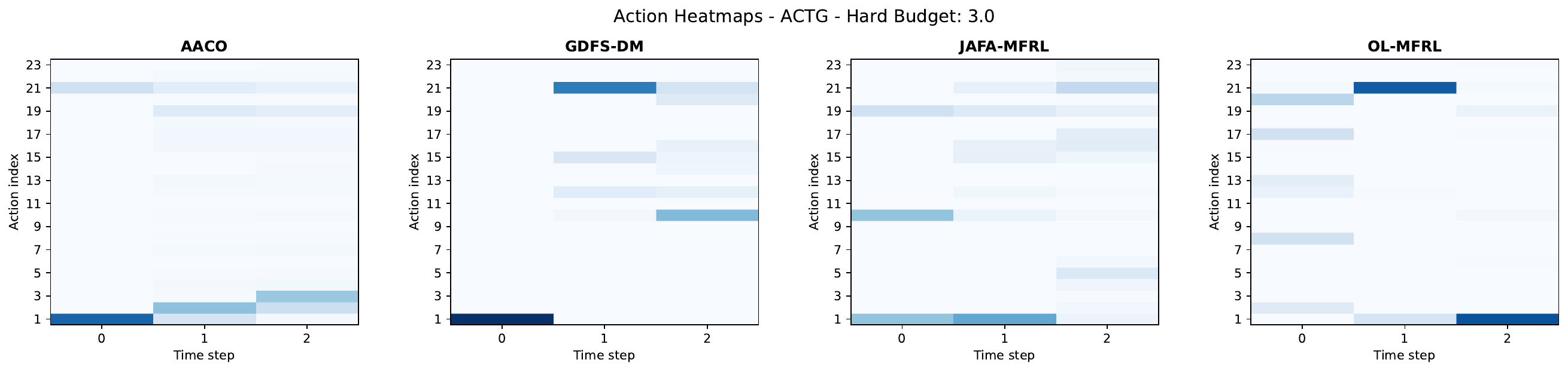}
	\caption{Heatmap of actions performed during evaluation with hard budget 3 on the \dataset{ACTG} dataset.}
	\label{fig:action_heatmap_actg_budget_3}
\end{figure*}

\begin{figure*}
	\centering
	\includegraphics[width=\textwidth]{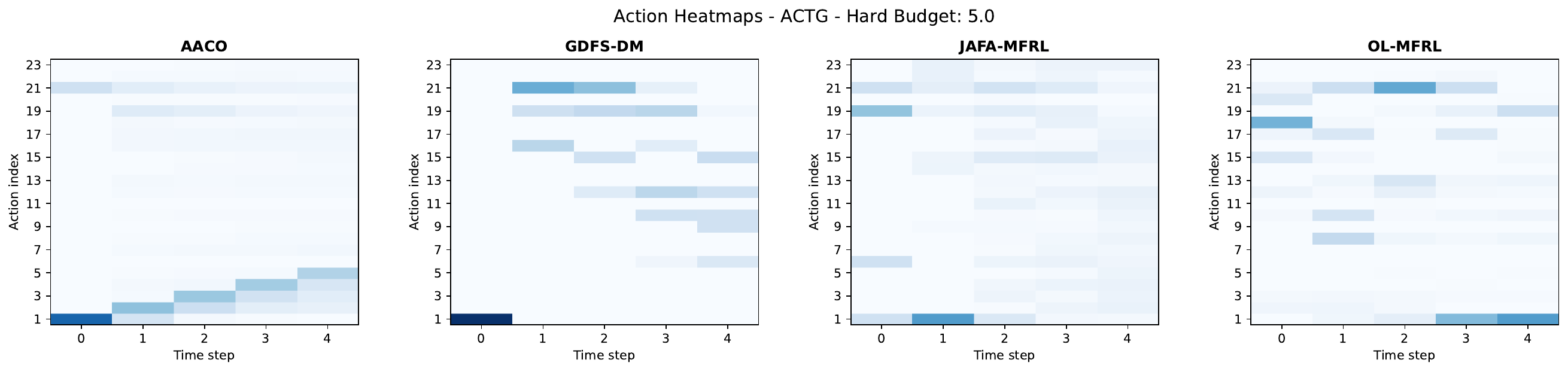}
	\caption{Heatmap of actions performed during evaluation with hard budget 5 on the \dataset{ACTG} dataset.}
	\label{fig:action_heatmap_actg_budget_5}
\end{figure*}

\begin{figure*}
	\centering
	\includegraphics[width=\textwidth]{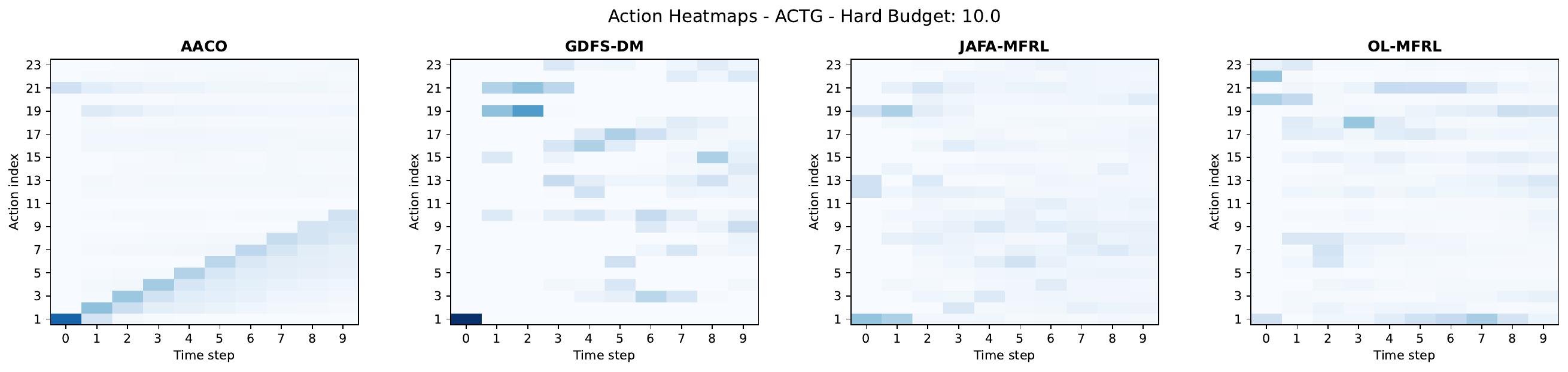}
	\caption{Heatmap of actions performed during evaluation with hard budget 10 on the \dataset{ACTG} dataset.}
	\label{fig:action_heatmap_actg_budget_10}
\end{figure*}

\Cref{fig:action_heatmap_actg_budget_3}, \Cref{fig:action_heatmap_actg_budget_5}, and \Cref{fig:action_heatmap_actg_budget_10} show heatmaps for actions performed on the \dataset{ACTG} dataset. We include this result due to the interesting behavior of \method{AACO}. It clearly learns a near-deterministic acquisition order, visible as a staircase pattern over time. This is consistent with the original \method{AACO} decision rule: first select a cost-aware subset, then, when multiple new features are suggested, execute only the single feature with lowest expected classification loss. In practice, this tie-breaking induces a stable feature ordering on \dataset{ACTG}.

\subsubsection{Action heatmaps on CUBE-NUC}

The last heatmaps we want to showcase are for the \dataset{CUBE-NUC} dataset. This dataset was included to test a method's ability to adapt to non-uniform costs. The last 10 features are identically distributed to the first 10 features, but cost twice as much. Hence, one would expect methods that can adapt to non-uniform costs to only choose the first 10 features. However, out of the four methods, \method{OL-MFRL} is the only method that consistently learns to do so across all budgets. \method{JAFA-MFRL} learns to do it for small budgets but fails for larger budgets. The original \method{GDFS-DM} does not account for feature acquisition costs. We incorporate costs by biasing the soft feature mask during training, but since costs do not directly influence the optimization objective, feature selection can still be affected by noise and biased towards costly features, indicating the limitation of this approach.

For \method{AACO}, this behavior follows from its two-stage action selection. The subset score includes acquisition cost, but when the chosen subset contains multiple unobserved features, the final single-feature tie-break is based on expected loss only \cite{pmlr-v235-valancius24a}. On \dataset{CUBE-NUC}, where each costly feature has a cheap counterpart with similar signal, this can still lead to selecting costly features, especially at larger budgets and with noisy $k$-NN density estimates.

\begin{figure*}
	\centering
	\includegraphics[width=\textwidth]{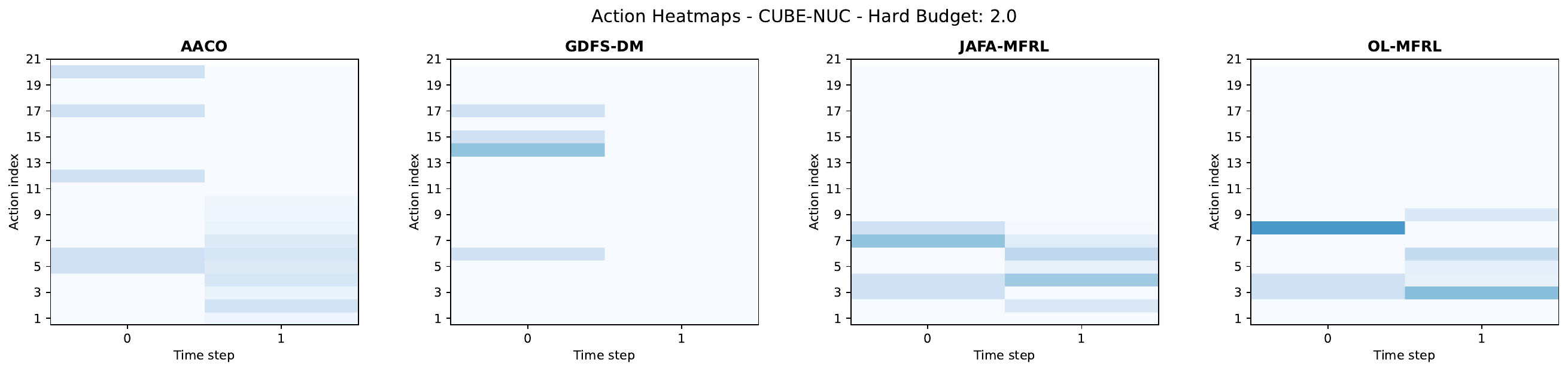}
	\caption{Heatmap of actions performed during evaluation with hard budget 2 on the \dataset{CUBE-NUC} dataset.}
	\label{fig:action_heatmap_cube_nuc_budget_2}
\end{figure*}

\begin{figure*}
	\centering
	\includegraphics[width=\textwidth]{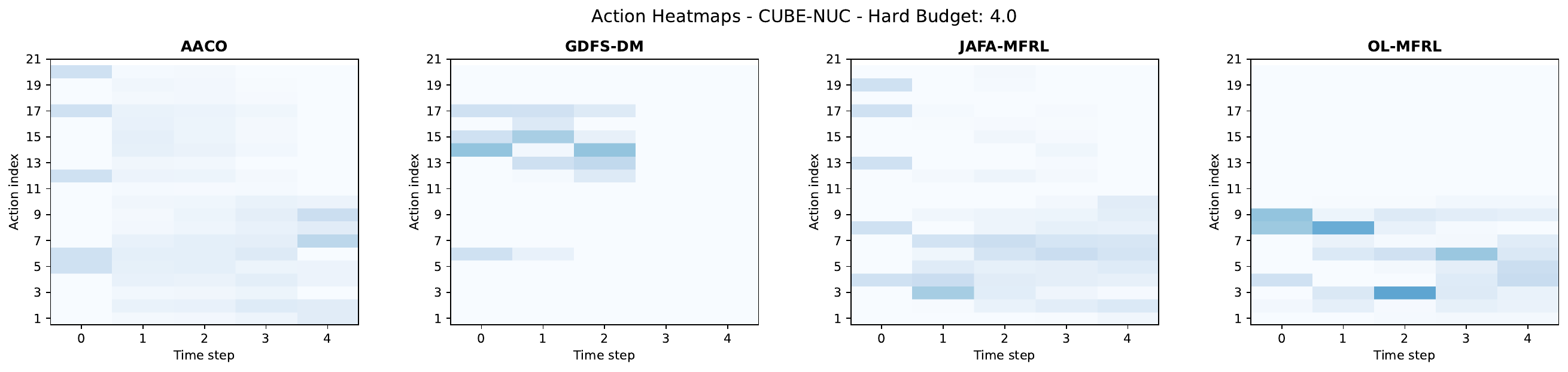}
	\caption{Heatmap of actions performed during evaluation with hard budget 4 on the \dataset{CUBE-NUC} dataset.}
	\label{fig:action_heatmap_cube_nuc_budget_4}
\end{figure*}

\begin{figure*}
	\centering
	\includegraphics[width=\textwidth]{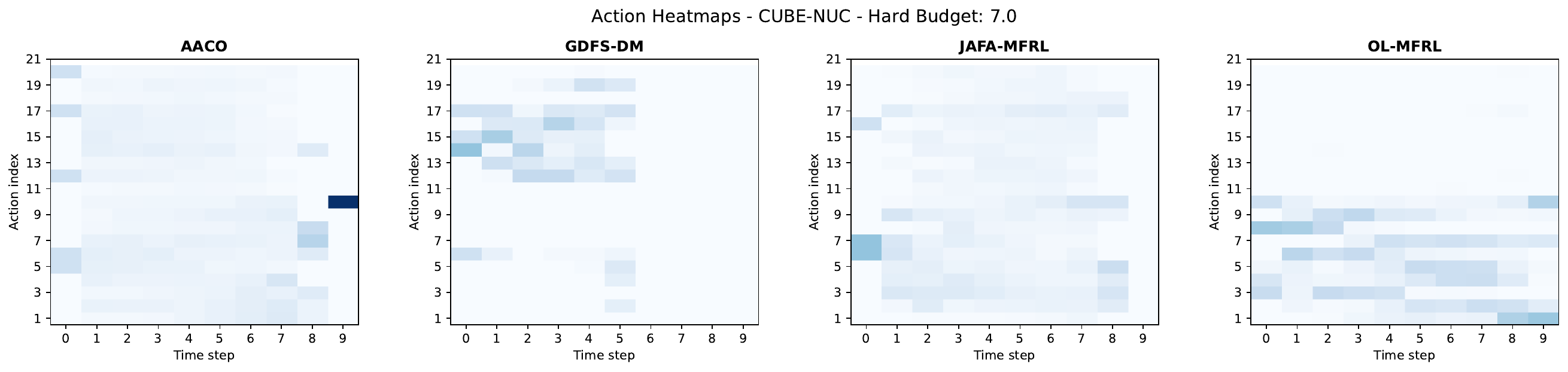}
	\caption{Heatmap of actions performed during evaluation with hard budget 7 on the \dataset{CUBE-NUC} dataset.}
	\label{fig:action_heatmap_cube_nuc_budget_7}
\end{figure*}

\subsection{Compute comparison} \label{appendix:compute}

\begin{figure*}
	\centering
	\includegraphics[width=\textwidth]{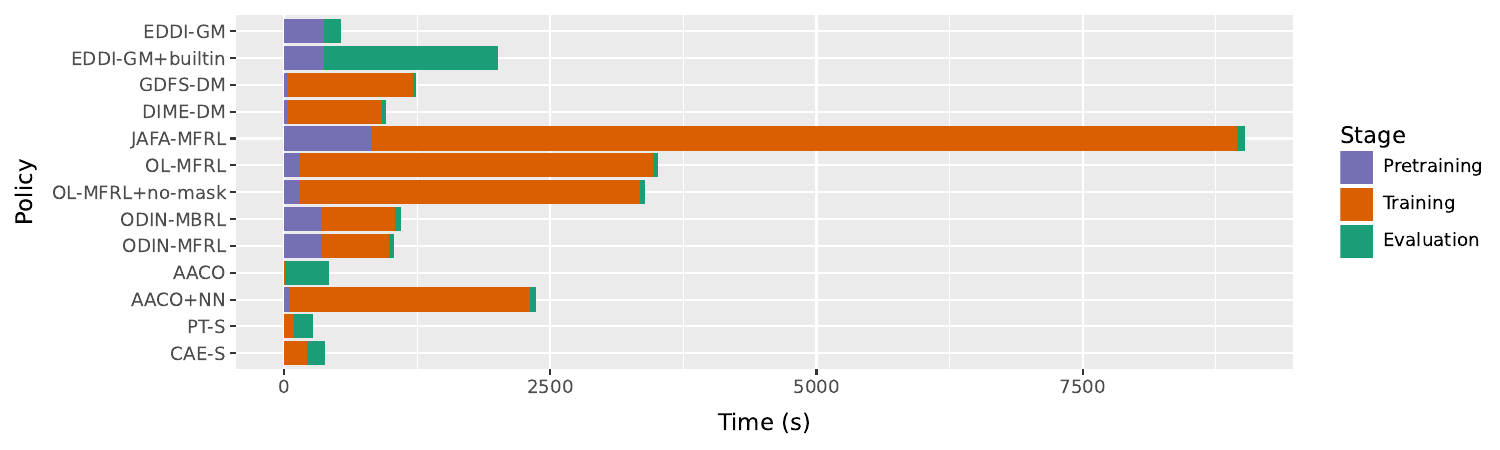}
	\caption{The average time required to pretrain, train, and evaluate each policy, averaged over common datasets and five seeds.}
	\label{fig:time}
\end{figure*}

\Cref{fig:time} shows the time required to pretrain, train, and evaluate each policy, averaged over only datasets that all methods were evaluated on. All methods used identical hardware, except for \method{AACO(+NN)} which was trained on CPU instead of GPU. Nevertheless, \method{JAFA} was the method that required the most compute, likely due to its use of recurrent neural networks.

The decomposition into pretraining, training, and evaluation also highlights that methods have different runtime bottlenecks. RL-based methods are generally training-heavy, while oracle-style methods shift more cost to evaluation through repeated candidate scoring. This distinction is important in practice: depending on whether one prioritizes faster development cycles or lower online inference latency, different methods may be preferable even at similar predictive performance.

\subsection{Method variations} \label{appendix:method_variations}

\subsubsection{ODIN}

\begin{figure*}
	\centering
	\includegraphics[width=\textwidth]{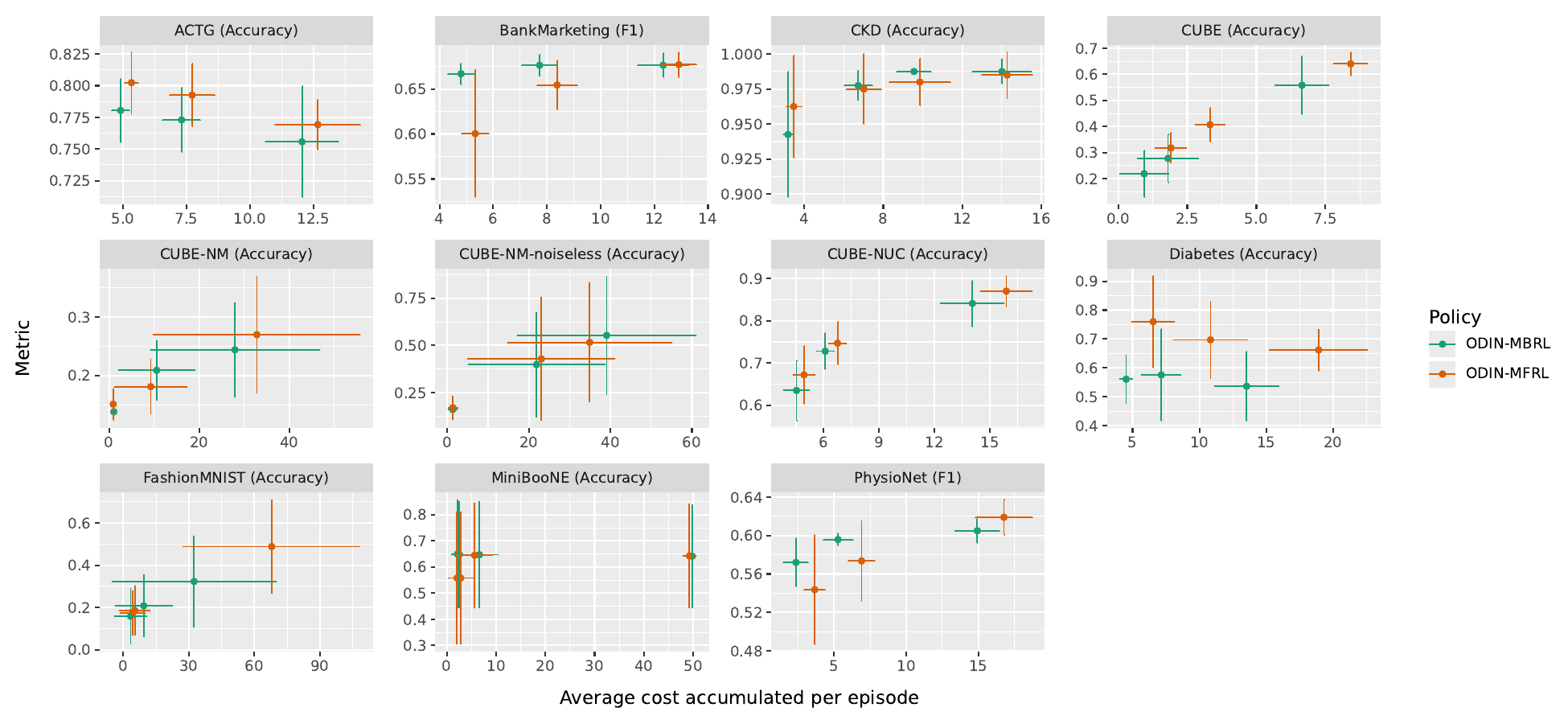}
	\caption{Soft-budget results using the built-in classifier of \method{ODIN}, showing the difference between the model-based and model-free variant.}
	\label{fig:odin_variants}
\end{figure*}

\Cref{fig:odin_variants} illustrates soft-budget results with the built-in classifier for the two \method{ODIN} variants. The methods behave similarly on most datasets, and with only minor differences. The model-based variant seems to perform better on \dataset{BankMarketing}, while the model-free version performs better on \dataset{ACTG} and \dataset{Diabetes}. The lack of a clear separation between the methods is not surprising, since the original paper \cite{zannone2019odin} found that the most significant difference could be observed for datasets with a size between 96 and 128. \dataset{CKD} is the smallest dataset we investigate, with $240$ training samples, for which \method{ODIN-MBRL} is slightly better, but within the margin of error.

\subsubsection{OL-MFRL}

\begin{figure*}
	\centering
	\includegraphics[width=\textwidth]{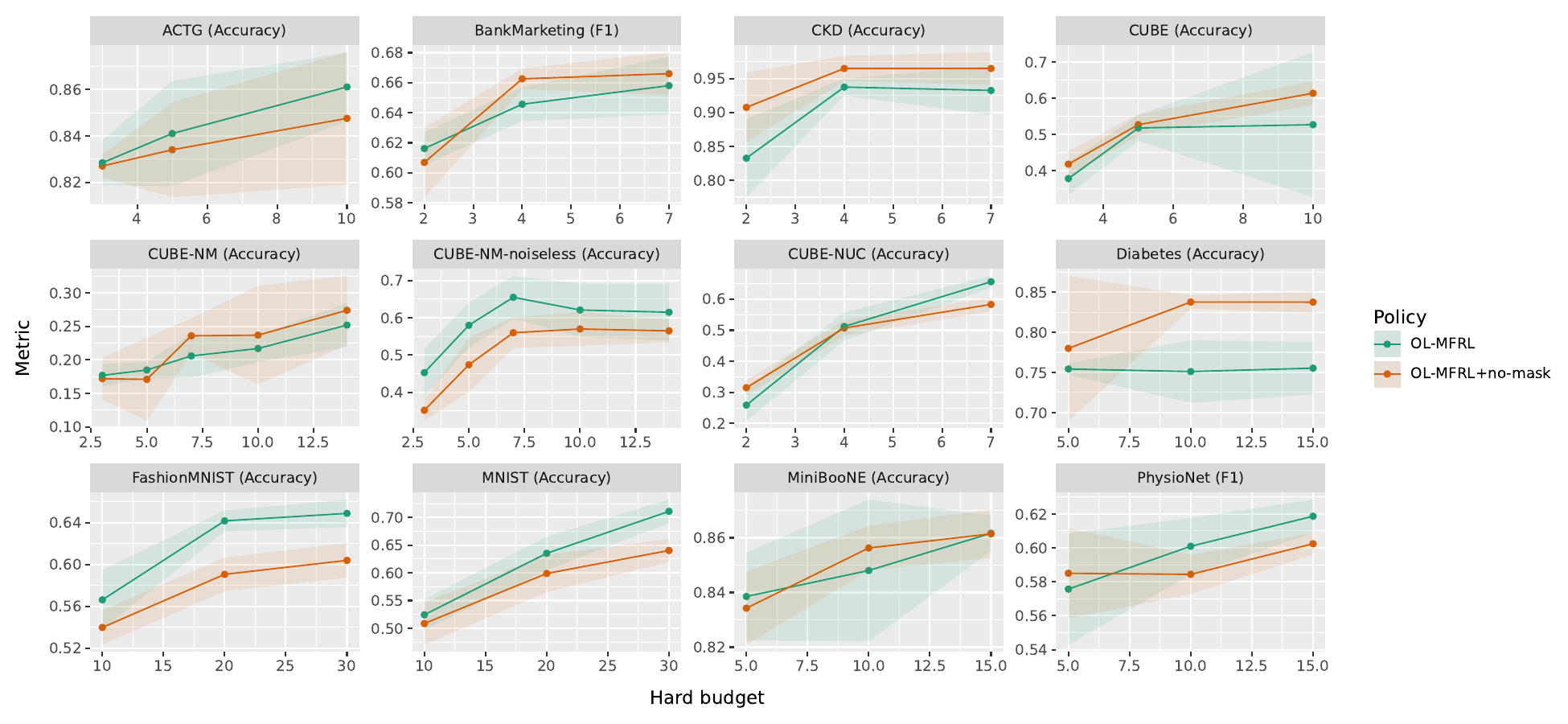}
	\caption{Hard-budget results using the built-in classifier of \method{OL-MFRL}, showing the impact of concatenating the feature mask to the input.}
	\label{fig:ol_variants}
\end{figure*}

\Cref{fig:ol_variants} shows hard-budget results with the built-in classifier for the two \method{OL-MFRL} variants. Adding feature mask information seems to be most useful on \dataset{CUBE-NM-noiseless} and \dataset{(Fashion)MNIST}. \dataset{CUBE-NM-noiseless} uses the value $0.0$ for both the context features and the three informative features encoding the class, while in \dataset{(Fashion)MNIST} the value $0.0$ represents black pixels.

As expected, \method{OL-MFRL+no\_mask} performs better on the \emph{noisy} version of \dataset{CUBE-NM}, where $0.0$ is a very unlikely feature value, making the feature mask redundant. Interestingly, \method{OL-MFRL+no\_mask} also performs substantially better on \dataset{Diabetes}.

\subsubsection{EDDI}
\begin{figure*}[t]
	\centering
	\includegraphics[width=\textwidth]{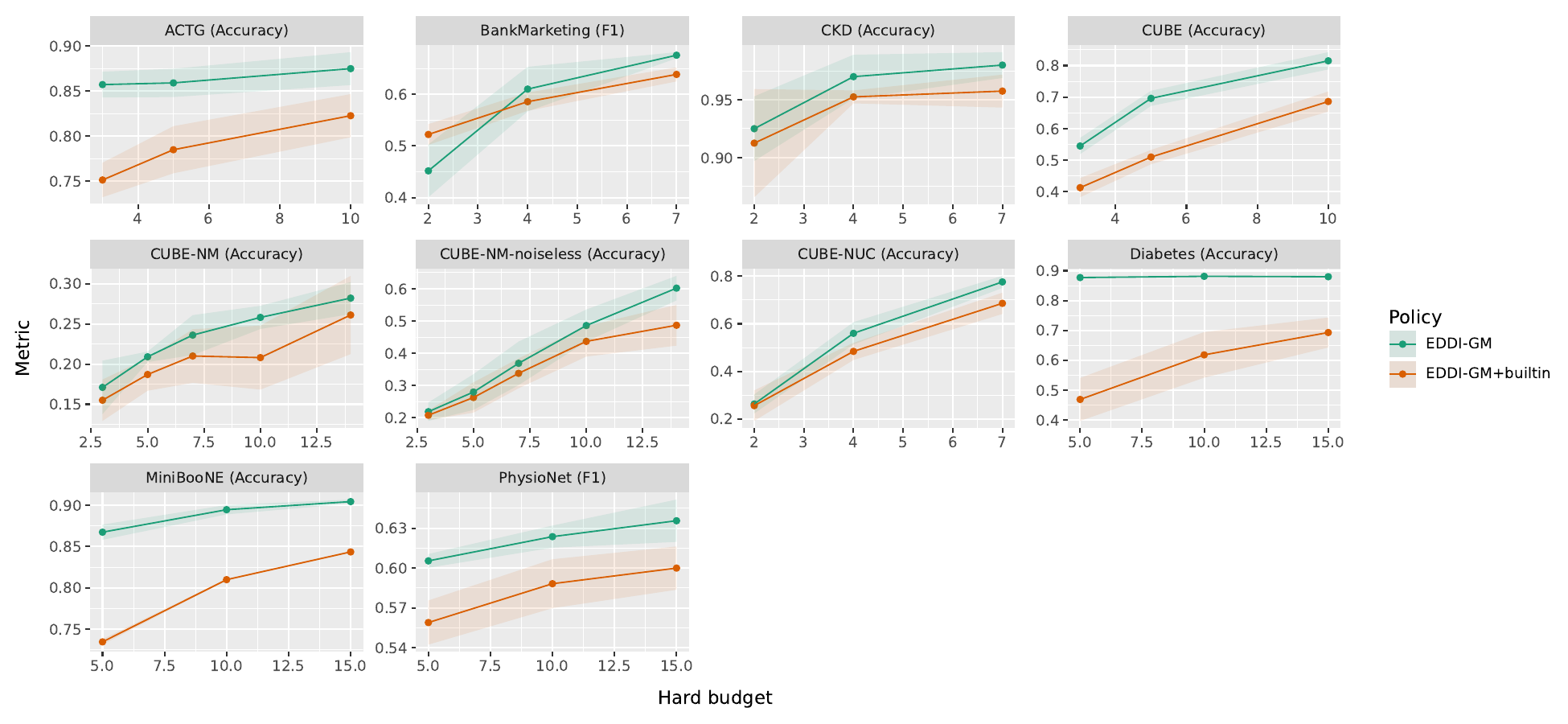}
	\caption{Hard-budget classification performance of two \method{EDDI} variants using external classifiers.}
	\label{fig:eddi_variants}
\end{figure*}

As shown in \Cref{fig:time}, since the \method{EDDI-GM} variant does not require two rounds of Monte Carlo sampling during feature selection, its inference time in the feature selection stage is significantly lower than that of the \method{EDDI-GM+builtin} variant, resulting in a much shorter evaluation time. From \Cref{fig:hard_budget_main_external_all} and \Cref{fig:hard_budget_main_builtin_all}, we can also observe that the built-in classifier of \method{EDDI} consistently achieves worse prediction performance than the external classifier. Moreover, \Cref{fig:eddi_variants} further shows that, when using the same external classifier for prediction, under the same hard-budget values, \method{EDDI-GM} with an external classifier for feature selection achieves better prediction performance in most cases compared to the \method{EDDI-GM+builtin} variant. These results indicate that using a more accurate external classifier during feature selection enables a more efficient and accurate assessment of feature informativeness, which in turn leads to the selection of more effective feature subsets under the same budget. Therefore, in the main experiments, we use the \method{EDDI-GM} variant with an external classifier to compare with other methods.

\subsubsection{AACO}
\begin{figure*}
	\centering
	\includegraphics[width=\textwidth]{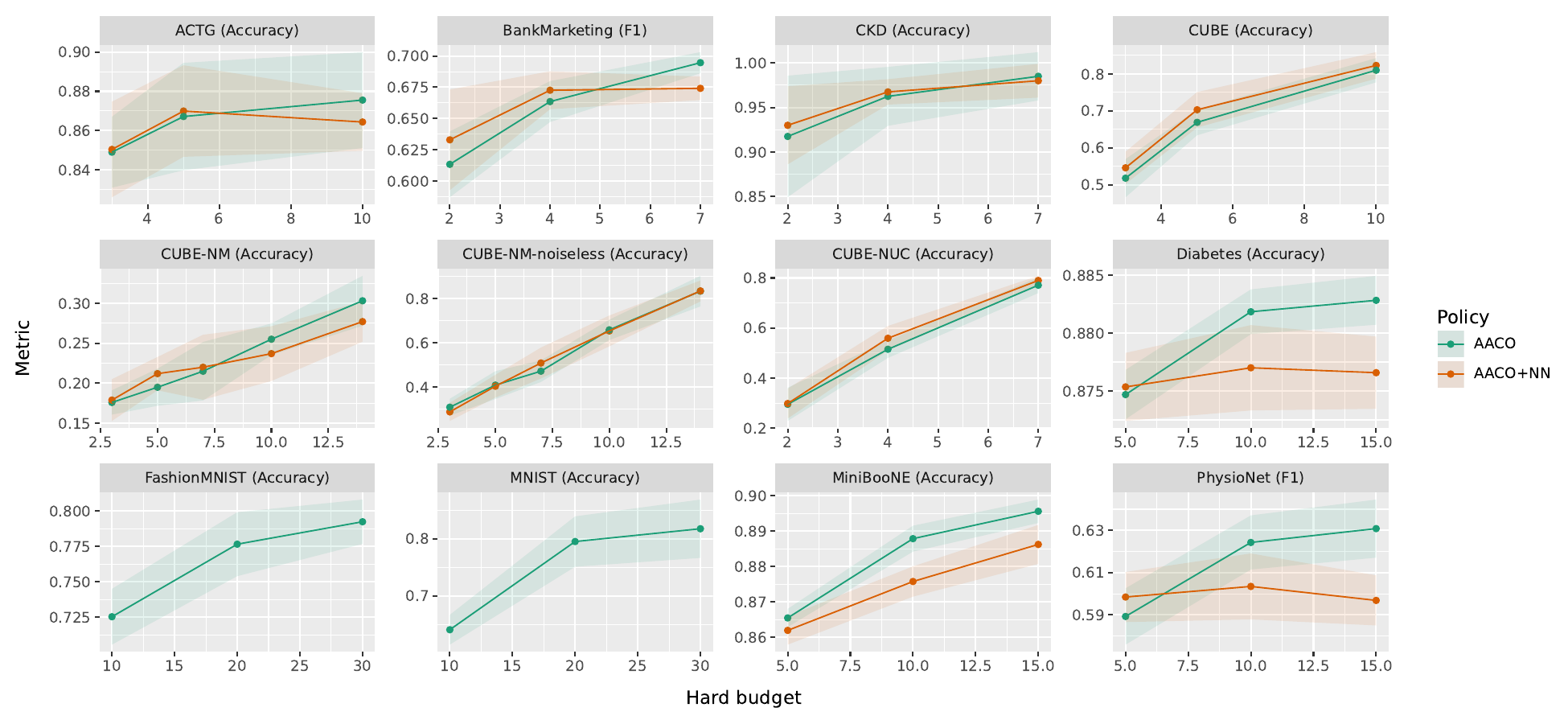}
	\caption{Hard-budget results using external classifiers for \method{AACO} and \method{AACO+NN}.}
	\label{fig:aaco_variants}
\end{figure*}

\Cref{fig:aaco_variants} compares \method{AACO} and \method{AACO+NN} in the hard-budget setting using external classifiers. \method{AACO+NN} is an imitation-learning approximation of \method{AACO}: it learns from oracle rollouts and then replaces the expensive $k$-NN subset search with a single neural forward pass at inference. The two methods are broadly consistent, but \method{AACO+NN} can deviate on some datasets due to approximation and compounding rollout errors. In return, \method{AACO+NN} is substantially cheaper to run at test time, see \Cref{fig:time}.

\subsubsection{Static Methods}

\begin{figure*}
	\centering
	\includegraphics[width=\textwidth]{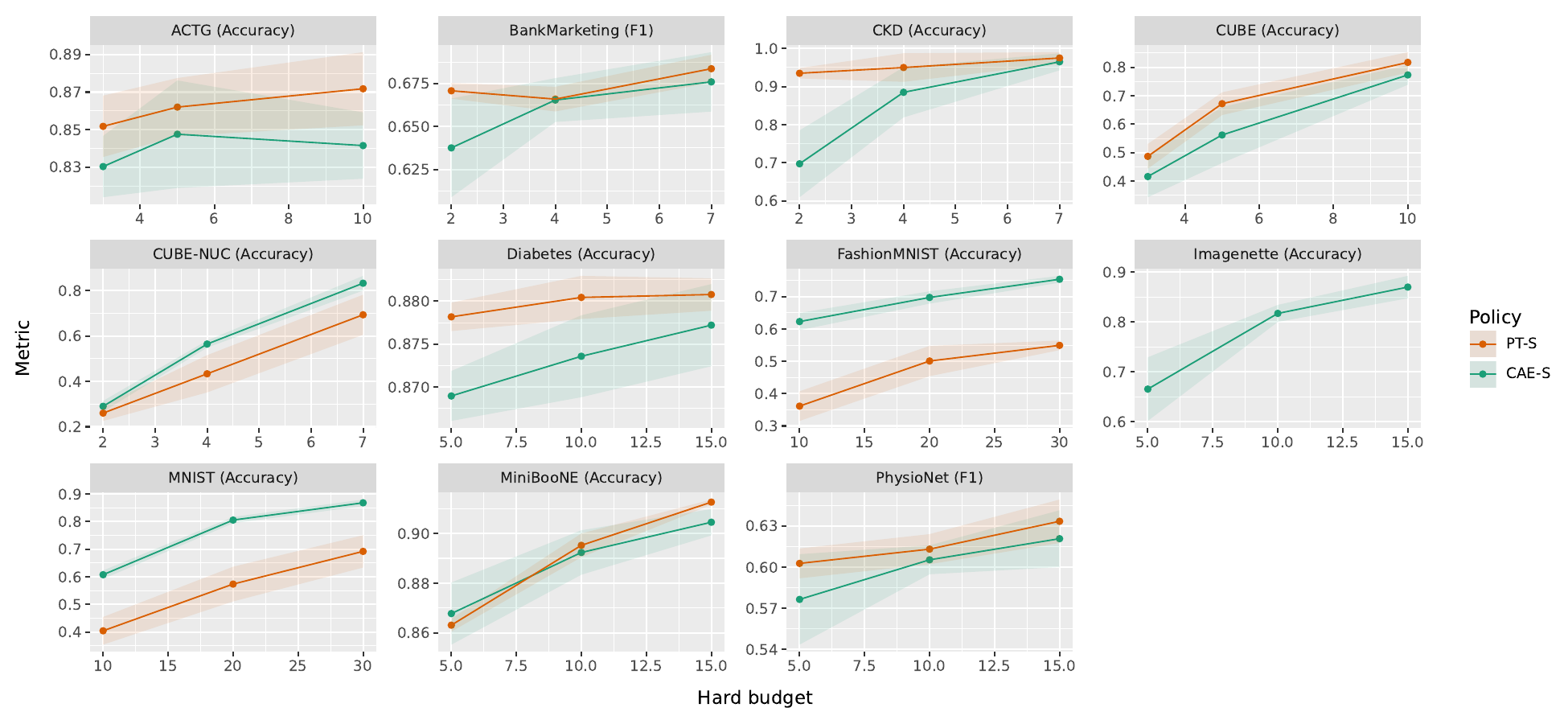}
	\caption{Hard-budget results using external classifiers for the two static methods \method{PT-S} and \method{CAE-S}.}
	\label{fig:static_variants}
\end{figure*}

\Cref{fig:static_variants} shows hard-budget results for two static feature selection methods, \method{PT-S} and \method{CAE-S}. They show similar performance on most datasets, except for \dataset{(Fashion)MNIST} where \method{CAE-S} is superior, and \dataset{CKD} where \method{PT-S} is better for smaller budgets. From these results, we concluded that \method{CAE-S} is a slightly stronger baseline, hence its inclusion in the main results.

\section{Ensuring a Fair Benchmark} \label{appendix:fair-benchmark}

In this section, we discuss the assumptions necessary to provide a fair and objective benchmark.

\textbf{Hard vs.\ soft budget}. As described in \Cref{section:scope}, we consider two special cases of \eqref{eq:afaopt}, referred to as the \emph{hard-budget} and \emph{soft-budget} settings. Most existing methods target only one of these settings, and comparisons are often confounded by including baselines from the other setting, since performance can differ substantially between hard and soft budgets. To ensure clarity and fairness, this benchmark reports results separately for (i) the soft-budget setting and (ii) the hard-budget setting. For each setting, methods are adapted when possible to match the corresponding objective. In \Cref{appendix:methods}, we describe how methods are adapted to both settings.

\textbf{Common components}. All methods are implemented in a unified framework, where common components between the methods are consistent. We now provide some examples of this. (i) Both \method{EDDI-GM} and \method{ODIN-MBRL} use a pre-trained partial vartiational autoencoder (although for different purposes). See \Cref{appendix:methods} for details on how it is used for the two methods. We use a pre-trained (external) predictor $f$ for each dataset, shared across all methods. However, some methods are designed to learn an (internal) predictor $f$ jointly with the policy $\pi$. For such methods, we report results for both the external predictor and the learned internal predictor (see next point).

\textbf{Internal vs.\ external predictor}. As discussed in \Cref{section:afaformulation,section:external_vs_internal_predictor}, some methods learn the predictor $f$ jointly with the acquisition policy $\pi$, whereas others assume that $f$ is trained independently. To enable fair comparisons, we therefore report results for two evaluation protocols: (i) all methods use the same shared (external) predictor, and (ii) methods that support joint training are additionally evaluated using their jointly trained (internal) predictor.

\textbf{Consistent RL framework}
Agents in RL-based methods use essentially the same training loop, only differing in which RL algorithm and reward function is used. See \Cref{appendix:methods} for details.

\textbf{Excluded methods}. As discussed in \citep{aronsson2025surveyactivefeatureacquisition}, there exist additional approaches to AFA beyond those included in this benchmark. We exclude these methods because they are either (i) tied to a specific predictor class (e.g., decision trees), or (ii) depend on strong assumptions about the data distribution (e.g., naive Bayes). In contrast, the benchmark focuses on general, model-agnostic methods that are compatible with a broad range of domains and predictors.

\textbf{Randomness}. In our experiments, we report the average performance over (i) multiple data splits (i.e., training, validation, and test). We also include the standard deviation across runs, visualized as error bars.

\textbf{Adherence to original architecture}
Although the most fair approach would be to perform extensive hyperparameter sweeps for each combination of methods/data sets/budgets, this is not realistic in practice. Therefore, we have opted to use the hyperparameters presented in the original paper / repository for each method, although some exceptions to this rule do exist. For example, when the parameters of a method have been tuned to datasets with a small number of features and we find that it underfits to \dataset{(Fashion)MNIST}, we have increased the complexity of the model. Some other changes have been made for consistency reasons. A list of the most important changes made compared to the original implementation can be found in Appendix \ref{appendix:experiments}.

\section{Detailed Description of Included Datasets} \label{appendix:datasets}

\begin{table*}[t]
\caption{Summary of datasets used in the benchmark.}
\label{tab:datasets}
\centering
\begin{tabular}{lcccccccc}
  \toprule
  Dataset & Type & Modality & Train Size & Val Size & Test Size &
  \# Features & \# Groups & \# Classes \\
  \midrule
  CUBE & Synthetic & Tabular & 600 & 200 & 200 & 20 & 20 & 8 \\
  CUBE-NUC & Synthetic & Tabular & 600 & 200 & 200 & 20 & 20 & 8 \\
  CUBE-NM & Synthetic & Tabular & 600 & 200 & 200 & 55 & 51 & 8 \\
  CUBE-NM-noiseless & Synthetic & Tabular & 600 & 200 & 200 & 55 & 51 & 8 \\
  MNIST & Real World & Image (tabularized) & 36,000 & 12,000 & 12,000 & 784 & 784 & 10 \\
  FashionMNIST & Real World & Image (tabularized) & 36,000 & 12,000 & 12,000 & 784 & 784 & 10 \\
  Diabetes & Real World & Tabular & 55,237 & 18,412 & 18,413 & 45 & 45 & 3 \\
  PhysioNet & Real World & Tabular & 7,200 & 2,400 & 2,400 & 41 & 41 & 2 \\
  MiniBooNE & Real World & Tabular & 78,038 & 26,012 & 26,014 & 50 & 50 & 2 \\
  ACTG175 & Real World & Tabular & 1,283 & 427 & 429 & 23 & 23 & 2 \\
  CKD & Real World & Tabular & 240 & 80 & 80 & 24 & 24 & 2 \\
  BankMarketing & Real World & Tabular & 27,126 & 9,042 & 9,043 & 16 & 16 & 2 \\
  Imagenette & Real World & Image & 5,681 & 3,788 & 3,925 & 150,528 & 196 & 10 \\
  \bottomrule
\end{tabular}
\end{table*}

\begin{figure*}[h]
	\centering
	\includegraphics[width=\textwidth]{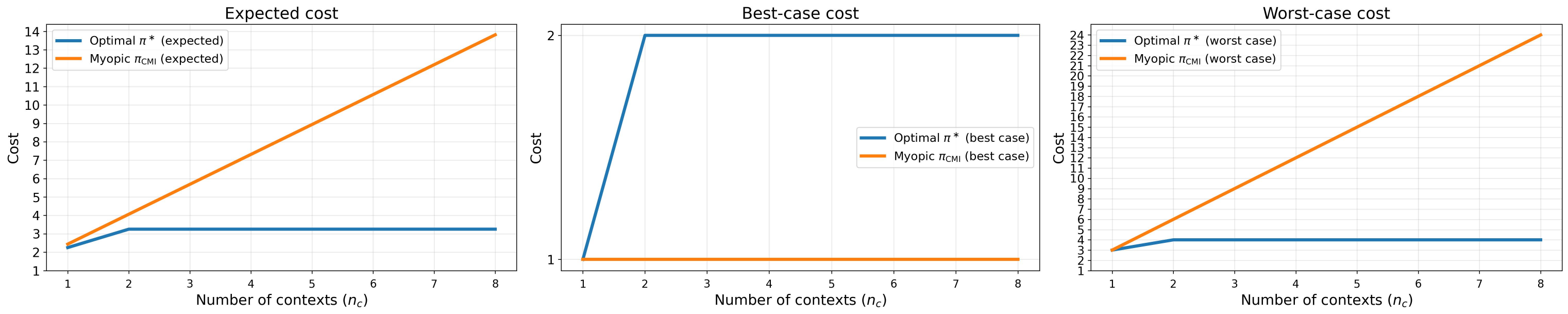}
	\caption{Query cost on noiseless \dataset{CUBE-NM} as a function of the number of contexts $n_c$. Left: expected cost $\mathbb{E}_{\mathbf{x}}\mathbb{E}_{\pi}[c(\pi[\mathbf{x}])]$. Middle: best-case cost $\min_{x\in\mathcal{X}} c(\pi[x])$. Right: worst-case cost $\max_{x\in\mathcal{X}} c(\pi[x])$. Curves compare the optimal policy $\pi^\ast$ and the myopic CMI policy $\pi_{\mathrm{CMI}}$, as characterized in Theorem~\ref{thm:cube-nm} (also see \Cref{appendix:proofcube}).}
	\label{fig:cube_nm_three_panels}
\end{figure*}

A summary of all datasets is provided in Table~\ref{tab:datasets}.
\dataset{CUBE} is a widely used synthetic dataset in the AFA literature, which we use to test each policy's ability to adapt to non-uniform costs with \dataset{CUBE-NUC}.
We describe its structure and limitations in Section~\ref{section:syntheticdata}, and introduce a new synthetic dataset, \dataset{CUBE-NM}, to address an important shortcoming of \dataset{CUBE}. Moreover, Figure \ref{fig:cube_nm_three_panels} illustrates the gap in performance between the myopic CMI policy in \eqref{eq:cmi} and an optimal policy, as $n_c$ increases (see Theorem \ref{thm:cube-nm}).

One important implementation detail is that the context variable denoted by $f_1$ was encoded as a one-hot vector in our experiments, which simplified learning. Using a custom \texttt{Unmasker}, we allowed policies to unmask the entire one-hot vector with a single AFA action.

\dataset{MNIST} is the standard handwritten digit classification dataset \citep{DBLP:journals/pieee/LeCunBBH98}. Following prior work in AFA \citep{gadgil2024estimating}, we treat \dataset{MNIST} as a tabular task by considering each pixel as a separate feature. \dataset{FashionMNIST} is a dataset of grayscale images of clothing items such as shirts, trousers, and shoes \citep{xiao2017/online}. It shares the same format and image dimensions as \dataset{MNIST}, and we similarly treat each pixel as a separate feature for tabular modeling. 

\dataset{Diabetes} is a real-world dataset for diabetes diagnosis, making it naturally suitable for AFA since different patients may require different examinations for an accurate diagnosis. This dataset has been widely used in previous AFA research \citep{kachuee2018opportunistic, covert2023learning}. As in previous studies, we define three classes: normal, pre-diabetes, and diabetes, based on standard fasting glucose thresholds. The data originates from the National Health and Nutrition Examination Survey (NHANES) \citep{NHANES2018}, a large-scale, ongoing health survey of U.S. adults and children. We use the pre-processed version made available by \citet{kachuee2018opportunistic}, which includes data collected between 1999 and 2016. 

\dataset{PhysioNet} is another medical dataset, derived from the PhysioNet Challenge 2012 \citep{Goldberger2000PhysioNet}. It contains electronic health records (EHR) from ICU patients, with the goal of predicting in-hospital mortality, a binary classification task, based on results from various clinical tests and measurements. 

\dataset{MiniBooNE} is a particle identification dataset collected by the MiniBooNE experiment at Fermilab \citep{miniboone_particle_identification_199}, where the task is to distinguish signal events (electron neutrinos) from background events (muon neutrinos) based on reconstructed particle interaction features. The dataset is fully tabular, each instance is represented by a set of real-valued detector-level features. 

\dataset{ACTG175} is a clinical trial dataset from the UCI repository \citep{ACTG175}. \dataset{CKD} (chronic kidney disease) and \dataset{BankMarketing} are standard UCI tabular datasets \citep{CKD,bank_marketing}. \dataset{Imagenette} is a 10-class subset of ImageNet used here to evaluate patch-based image acquisition \citep{Imagenette}.


\section{Detailed Description of Implemented AFA Methods} \label{appendix:methods}

\begin{table*}[t]
  \centering
  \caption{Summary of feature selection methods included in the benchmark. The possible categories include \emph{generative myopic} (GM), \emph{discriminative myopic} (DM), \emph{model-free RL} (MFRL), \emph{model-based RL} (MBRL), and \emph{static selection} (S). 
  }
  \label{tab:feature-taxonomy}

  \renewcommand{\arraystretch}{1.2}
  \setlength{\tabcolsep}{5pt}

  \begin{tabularx}{\linewidth}{|
        >{\centering\arraybackslash}m{4.2cm}|   
        >{\raggedright\arraybackslash}m{4.5cm}| 
        >{\centering\arraybackslash}m{1.7cm}|   
        >{\raggedright\arraybackslash}X|}       
    \hline
    \textbf{Paradigm} & \textbf{Strategy} & \textbf{Myopic?} & \textbf{Representative Method(s)} \\ \hline

    \multirow[c]{4}{=}{\\AFA\\(Dynamic Feature Selection)} &
      Generative estimation of CMI & Myopic &
      EDDI-GM \citep{ma2019eddi} \\ \cline{2-4}

    & Discriminative estimation of CMI & Myopic &
      GDFS-DM \citep{covert2023learning} \newline
      DIME-DM \citep{gadgil2024estimating} \\ \cline{2-4}

    & Model-free RL & Non-myopic &
      JAFA-MFRL \citep{NEURIPS2018_e5841df2} \newline
      OL-MFRL \citep{kachuee2018opportunistic} \newline
      ODIN-MFRL \citep{zannone2019odin} \\ \cline{2-4}

    & Model-based RL & Non-myopic &
      ODIN-MBRL \citep{zannone2019odin} \\ \cline{2-4}

    & Oracle-based & Non-myopic &
      AACO \citep{pmlr-v235-valancius24a} \\ \hline

    \multirow[c]{1}{=}{Static Feature Selection} &
      Global feature importance & N/A &
      PT-S \citep{DBLP:journals/ml/Breiman01} (Permutation tests)\newline
      CAE-S \citep{DBLP:conf/icml/BalinAZ19} \\
      \hline
  \end{tabularx}
\end{table*}

As described in the recent survey by \citet{aronsson2025surveyactivefeatureacquisition}, AFA methods can be broadly categorized as follows:
(i) \emph{myopic} heuristics, which perform no explicit planning beyond a one-step lookahead and are often further classified as \emph{discriminative} or \emph{generative},
(ii) \emph{non-myopic} methods based on reinforcement learning (RL), which learn acquisition policies by exploiting the MDP structure underlying AFA (see \Cref{section:mdpformulation}) and can be divided into \emph{model-free} and \emph{model-based} approaches, and
(iii) \emph{non-myopic} methods that do not rely on explicit RL.

In this benchmark, we have selected at least one representative state-of-the-art method from each category to provide a comprehensive evaluation of the various strategies for AFA. Table~\ref{tab:feature-taxonomy} summarizes the methods included in the benchmark and their corresponding categories. In addition, we include two static feature selection methods to highlight the potential benefits of dynamic selection.

Below, we briefly describe each method and explain how it is adapted to the hard- and soft-budget settings. 
In the hard-budget setting, we use the same budget $B$ for all methods. 
In the soft-budget setting, we tune the threshold parameter $\alpha$ separately for each method, since it is implemented and interpreted slightly differently across methods (see the discussion below).

\subsection{Myopic methods}


All myopic methods considered in this benchmark are based on the myopic CMI policy in \eqref{eq:cmi} (i.e., $I(\mathbf{y}; \mathbf{x}_i \mid x_S)$). All methods in this category hence differ in how they compute this information gain. At each decision point, computing $I(\mathbf{y}; \mathbf{x}_i \mid x_S)$ for all feature indices $i$ requires access to the joint and conditional distributions of the features and the target, conditioned on the observed subset. In the following, we describe how the included myopic methods achieve this and how each are adapted to the hard/soft-budget setting.

For all myopic methods, unless stated otherwise, we adapt them to the hard- and soft-budget settings as follows. 
Under the hard-budget setting, the adaptation is immediate: we iteratively acquire features according to \eqref{eq:cmi} until the total cost reaches the budget $B$. 
Under the soft-budget setting, we adopt a commonly used heuristic and continue acquiring features via \eqref{eq:cmi} until it falls below a threshold for every remaining feature, that is, until $I(\mathbf{y};\mathbf{x}_i \mid \mathbf{x}_S)/c_i < \alpha$ for all $i \in [d]\setminus S$ \citep{gadgil2024estimating, aronsson2025surveyactivefeatureacquisition}. To ensure consistency with the other methods in the benchmark, we slightly deviate from the original experimental setting for the discriminative myopic methods. For datasets with uniform costs, under the hard-budget setting, we use the same hard budget values for both training and evaluation, and under the soft-budget setting, we train these methods with the maximum training hard budget and then reuse them across different soft-budget thresholds. For datasets with non-uniform costs, we train the discriminative myopic methods with the maximum hard budget (i.e., all features) to avoid acquiring substantially more features during evaluation than during training.

\textbf{EDDI-GM} \citep{ma2019eddi}. \method{EDDI} estimates the CMI acquisition policy using a \emph{generative} approach. At its core lies a novel partial Variational Autoencoder (PVAE) capable of handling arbitrarily missing features. Once trained on the full dataset $\mathcal{D}$, the PVAE offers fast approximations of posterior predictive distributions such as $p(\mathbf{x}_i \mid x_S)$: it samples a latent code conditioned on the observed subset $x_S$ and subsequently generates the candidate feature $\mathbf{x}_i$. Whenever these predictive distributions admit a Gaussian parameterization, the CMI can be computed analytically; otherwise, it is efficiently estimated via Monte Carlo sampling. A limitation of this approach is that the PVAE is expensive to train. This limitation is addressed in the next two \emph{discriminative} approaches.

Adapting EDDI-GM to the hard-budget setting is straightforward, by simply acquiring features according to \eqref{eq:cmi} until budget $B$ is reached. For the soft-budget setting, we use the following, commonly used heuristic: we acquire features according to $I(\mathbf{y};\mathbf{x}_i \mid x_S)/x_i$ until $I(\mathbf{y};\mathbf{x}_i \mid x_S)/x_i < \alpha$ for all $i \in [d] \setminus S$.   

\textbf{GDFS-DM} \citep{covert2023learning}. This approach formulates an optimization problem that evaluates a policy by its one-step-ahead prediction accuracy. Specifically, when the policy has chosen the next feature $i$ given the currently observed set $x_{S}$, the goal is to minimize the expected loss $\mathbb{E} \left[\,l\bigl(f(x_{S},\mathbf{x}_i),\mathbf{y}\bigr)\right]$, where $l$ measures the discrepancy between the prediction and the target $\mathbf{y}$. The optimal predictor for this objective is the Bayes classifier, $f^{*}(x_{S}) = p(\mathbf{y}\mid x_{S})$. Moreover, under this classifier, the policy that minimizes the expected one-step-ahead loss coincides with the CMI policy of Eq. \eqref{eq:cmi}. The resulting optimization is solved with amortized optimization \citep{amor2022}: the variational objective is expressed in a closed form and then minimized using a deep neural network. 

Since GDFS-DM simply outputs the action $a \in [d] \setminus S$ that would yield largest information gain (and not the actual information gain itself), we can not use the heuristic discussed earlier for the soft-budget case. Instead, we stop when $H(\mathbf{y} \mid x_S) < \alpha$ (based on $p(\mathbf{y} \mid x_S)$ output by predictor $f(S,x_S)$).

\textbf{DIME-DM} \citep{gadgil2024estimating}. This approach directly estimates the CMI in a discriminative way. Extending the variational estimators of \citep{chattopadhyay2023variational,covert2023learning}, introduces a new objective whose optimum equals the true CMI. The practical implementation uses two jointly trained networks: a predictor $f_\theta$ and a value network $g_\phi$ that outputs $I(\mathbf{y}; \mathbf{x}_i \mid x_S)$ for each unobserved feature. When $f_\theta$ approaches the Bayes classifier, the expected reduction in cross-entropy achieved by adding the feature $i$ matches its CMI; training $g_\phi$ to reproduce this reduction yields a consistent estimator. After training, $g_\phi$ selects the next feature by choosing the largest predicted CMI.

\subsection{Non-myopic methods via reinforcement learning}

As discussed in \Cref{section:mdpformulation}, AFA can be formulated as a Markov decision process (MDP) \citep{dulac2011datum, aronsson2025surveyactivefeatureacquisition}. Consequently, one can use reinforcement learning (RL) to find non-myopic policies $\pi$ that acquire feature that maximize long-term reward. This approach enables the discovery of policies that can outperform the myopic methods discussed above. Below, we describe all RL-based approaches used in the benchmark.

Unless otherwise stated, all RL-based methods are adapted to the hard and soft-budget settings as follows. All methods are adapted to the hard-budget setting by simply setting $\alpha = 0$ in the reward (See \Cref{section:mdpformulation}). For the soft-budget setting, we set $\alpha > 0$, and set the budget $B = c([d])$ (cost of acquiring all features).

\textbf{JAFA-MFRL} \citep{NEURIPS2018_e5841df2}. This method uses the RNN-based and order-invariant set encoding introduced in \citep{vinyalsOrderMattersSequence2016} to represent the subsets of acquired features, significantly reducing the complexity of the search space. The approach jointly trains an RL agent and a classifier, ensuring that the feature selection policy is optimized with respect to the classification objective. To solve the underlying MDP, the method uses deep Q-Networks (DQN) \citep{mnih2015human}, enabling efficient policy learning even in high-dimensional feature spaces. The reward function is exactly as in \Cref{section:mdpformulation}. 


%
%



\textbf{ODIN-MBRL} \citep{zannone2019odin}. This is a \emph{model-based RL} framework for AFA. \method{ODIN} uses a pre-trained PVAE \citep{ma2019eddi} to approximate the conditional distribution of unobserved features $p(\mathbf{x}_U \mid x_S)$ for $U = [d] \setminus S$. Since the MDP transition dynamics can be derived from this distribution, \method{ODIN} performs model-based rollouts to simulate feature acquisition trajectories, improving data efficiency and robustness to missing features. In our experiments, we performed rollouts to augment the training set by $200\%$. 

In contrast to \method{JAFA-MFRL}, this method uses a \textit{dense} reward function. Let $y_{\text{true}}$ be the true class of instance $x$ and $p_f(y_{\text{true}} \mid x_S)$ be the probability of this class given by the predictor $f(x_S)$. Then, for any state $(S,x_S)$ and action $a \in \mathcal{A}_B(S)$
\begin{equation}
    r((S,x_S), a) = p_f(y_{\text{true}} \mid x_S, x_a) - \alpha c(S \cup \{a\})
\end{equation}
which motivates the agent to select the most informative features first. This can alleviate issues with sparse rewards, but may bias the method to more myopic policies (see Section \ref{section:mdpformulation} for more discussion).

The policy is trained using Proximal Policy Optimization (PPO) \citep{DBLP:journals/corr/SchulmanWDRK17}.

\textbf{ODIN-MFRL}
The key strength of \method{ODIN-MBRL} is its performance on smaller datasets \citep{zannone2019odin}. However, since most of the datasets in this benchmark are relatively large, we decided to also evaluate a variant of \method{ODIN} where the agent is trained only on the original dataset, without using generative rollouts.

\textbf{OL-MFRL+no-mask} \citep{kachuee2018opportunistic}. This is another model-free and DQN-based method. The main characteristic of it is the reward function. For any state $(S,x_S)$ and action $a \in \mathcal{A}_B(S)$, the reward is defined as
\begin{equation}
    r((S,x_S), a) = \frac{\| \textit{Cert}(x_{S\cup a}) -  \textit{Cert}(x_S) \|}{c(a)}
\end{equation}
which does not depend on the true label but only on the classifier predictions. Here \textit{Cert} is a vector of class probabilities, calculated by averaging over multiple forward passes with dropout. Furthermore, the method uses a coupled PQ-network where the output of the predictor (P-network) is fed into the action-value estimator (Q-network).

This method was originally introduced for the hard-budget setting. Although it effectively adapts to non-uniform costs due to the denominator, every reward is non-negative, providing no incentive for the agent to stop early. Consequently, adapting it to the
soft-budget setting would require substantial changes to the reward design and we therefore omit this method from the soft-budget experiments. Our goal is to benchmark existing AFA methods as they are, and avoid major modifications that could reasonably be viewed as defining a different method.

\textbf{OL-MFRL}
One major drawback with \method{OL-MFRL+no-mask} is that it cannot distinguish between missing features and features with a value of $0$. Hence, \textbf{OL-MFRL} is a variation of the original \textbf{OL-MFRL+no-mask} method where we simply concatenate the feature mask to the features in the input.

\subsection{Oracle-based methods (non-myopic)}
Rather than learning policies through reinforcement learning, oracle-based methods directly approximate an optimal acquisition policy by estimating the underlying data distributions. This approach offers a middle ground between myopic methods and computationally expensive RL training.

\textbf{AACO} \citep{pmlr-v235-valancius24a} formulates AFA as a subset optimization problem. Instead of greedily selecting features one by one, ACO considers acquiring entire subsets of features simultaneously. The optimal policy selects the subset $o' \subseteq U$ that minimizes the expected acquisition cost plus the prediction loss,
\begin{equation}
u(x_o, o) = \argmin_{o' \subseteq U} \mathbb{E}_{y,x_{o'} \mid x_o}\left[\ell\left(\hat{y}(x_o, x_{o'}), y\right)\right] + \alpha |o'|,
\end{equation}
where $x_o$ denotes the currently observed features, $\mathcal{O}$ is the set of unobserved feature indices, $\hat{y}(x_o, x_{o'})$ is the prediction based on observed and newly acquired features, and $\alpha$ controls the acquisition cost per feature.

In practice, computing this expectation requires knowledge of the joint distribution $p(y, x_{o'} \mid x_o)$, which is unknown. \method{AACO} addresses this by using $k$-nearest neighbors density estimation to approximate the required distributions. Furthermore, for computational tractability in high-dimensional spaces, \method{AACO} samples a random subset of possible feature combinations rather than evaluating all $2^{|\mathcal{O}|}$ possibilities.

This approach is non-myopic since it jointly optimizes over multiple features rather than selecting them sequentially, yet it avoids the training complexity of RL methods by directly approximating the optimal oracle policy.

The method naturally supports the soft-budget setting by terminating early whenever $u(x_o,o)=\emptyset$, that is, when stopping and predicting with the currently acquired features achieves the highest utility. In the hard-budget setting, early termination is disallowed, and the method continues selecting features until the budget $B$ is exhausted.

\textbf{AACO+NN}
\method{AACO+NN} is a neural approximation of \method{AACO}. We first generate expert trajectories from a trained \method{AACO} oracle. Each trajectory contains states represented by
masked features and feature masks, and actions represented by either "stop" or "acquire selection $j$". We then train a policy network with behavioral cloning to imitate these oracle
actions. At inference time, this policy replaces repeated $k$-NN subset evaluation, while predictions are still made by the same classifier as in \method{AACO}.

\subsection{Static (non-adaptive) feature selection methods}

Static feature selection methods choose the same subset of features for all test instances, regardless of individual characteristics. We include them as baselines to highlight the potential advantage of AFA methods, which dynamically select features tailored to each instance. 

Static methods are restricted to the hard-budget setting. The reason is that they do not perform instance-wise selection, which makes the soft-budget setting irrelevant.

\textbf{PT-S} \citep{DBLP:journals/ml/Breiman01}. This method involves randomly permuting the values of a single feature in all data instances and measuring the resulting drop in model performance. By disrupting the relationship between the feature and the target, this approach reveals how much the model depends on that feature. When performing permutation tests, we evaluate validation accuracy by replacing the values in each feature column with random samples drawn from the corresponding column in the training set.

\textbf{CAE-S} \citep{DBLP:conf/icml/BalinAZ19}. This method is an end-to-end differentiable approach for global feature selection, designed to efficiently identify a subset of the most informative features while jointly training a neural network to reconstruct the input from the selected subset. Originally proposed for unsupervised feature selection via input reconstruction, we adapt it to supervised learning by modifying the prediction target, following \cite{covert2023learning}.

To adapt static feature selection methods to the active feature acquisition setting considered in this paper, we first identify a fixed set of the most informative features in advance on the training data. During evaluation, static methods use the same feature subset for all test instances under a hard budget $b$. For CAE-S, since it does not provide a relative ranking of feature importance, we use the CAE selector only to identify the optimal feature subset for the maximum budget, and during evaluation, features from this subset are selected sequentially in a fixed index-based order until the budget is exhausted. In contrast, PT-S selects the features sequentially in order of the estimated importance. This setup allows for a fair comparison with dynamic methods, which naturally select features one at a time.

\section{Experimental Details} \label{appendix:experiments}

\subsection{Implementation Details} 

\subsubsection{External classifiers}

Two different external classifiers were trained. They both followed a standard supervised learning loop, with the exception of using masking on each batch retrieved from the dataset. For each batch, every feature was masked uniformly with some probability that was drawn from a uniform distribution. The range of this distribution depended on the classifier and dataset. Checkpointing was used to save the model with lowest validation loss on batches with minimum masking probability.

The first classifier, which was used on all datasets except \dataset{Imagenette}, was a multilayer perceptron with the following common properties shared across datasets:
\begin{itemize}
    \item Learning rate: $1 \times 10^{-3}$
    \item Two hidden layers, each with $128$ units and ReLU activations.
\end{itemize}

Other properties, like batch size, number of training epochs, and masking probability, were slightly tuned for each dataset, to ensure good classification performance.

The second classifier, which was used on \dataset{Imagenette}, was a Vision Transformer with the following configuration:
\begin{itemize}
    \item Backbone: pretrained ViT (\texttt{vit\_small\_patch16\_224})
    \item Input resolution: $224 \times 224$
    \item Patch size: $16$
    \item Batch size: $128$
    \item Optimizer: Adam
    \item Training epochs: $200$
    \item Masking probability sampled from $\mathcal{U}(0,0.9)$
    \item Learning rate scheduling: ReduceLROnPlateau applied on the validation loss with a reduction factor of $0.2$, patience of $3$, initial learning rate of $1\times10^{-5}$, and minimum learning rate of $1\times10^{-8}$
\end{itemize}


\subsubsection{Method implementations}

Below, we aim to fully disclose how our method implementations differ from the original implementations. Although some methods do provide original implementations, they are often poorly documented and not compatible with PyTorch \citep{paszke2019pytorchimperativestylehighperformance}, which our framework is written in. In such cases, we have been forced to rewrite the method from scratch, verifying that the method works as expected on simple toy datasets.

In general, hyperparameters were always set to the values found in the paper or original implementation. In cases where a hyperparameter was not listed publicly, ''standard'' values have been chosen. Examples of this include setting the learning rate to $0.001$ and $\tau=0.005$ when soft-updating target networks for RL.

\subsubsection{Common across all methods}

\paragraph*{\textbf{Consistent masking probabilities during pretraining}}
Several methods needed a pretraining stage where the features are masked with varying probabilities. In our pretraining runs, we always used a fixed masking probability per batch of data, and the probability itself was sampled from the distribution
\begin{itemize}
    \item $\mathcal{U}(0.75,0.99)$ for \dataset{MNIST} and \dataset{FashionMNIST}
    \item $\mathcal{U}(0,0.9)$ for remaining datasets
\end{itemize}

Masking \textit{all} features is not useful, and since many methods can achieve reasonable performance on only very small percentage of the total number of features in (Fashion)MNIST, we increased the masking probability.

\textbf{Consistent loss function}
For methods that required a loss function, we used \emph{weighted} cross-entropy, using class probabilities from the training set.

\subsubsection{Common across RL methods}

TorchRL \citep{bou2023torchrl} was used extensively for the RL part. All agents were written using off the shelf loss modules, minimizing the risk for errors in the RL algorithms (like DQN and PPO). Moreover, a common AFA MDP environment was written, which all RL methods interacted with. 

Computations were performed on the GPU as much as possible. This means that both the agent (with replay buffers) and the environment were placed on the GPU. One exception to this was \method{OL-MFRL} which required the replay buffer to be placed in RAM when training on \dataset{(Fashion)MNIST}.

\textbf{Pretraining until convergence}
All of the RL methods have a pretraining stage, where a model is trained using supervised learning. Pretraining was always performed until convergence by using early stopping, always saving the model with the lowest validation loss on features masked with the minimum masking probability. As listed above, for (Fashion)MNIST this is 25\% and for the other datasets it is 10\%.

\textbf{No discounting}
We did not use discounting (therefore setting $\gamma = 1$) since the MDP is episodic.

\textbf{Consistent batch size, number of agents, and training epochs}
During the training stage, data was collected in consistent batches and with the same number of parallel agents for all RL methods. ``Agents'' in this case does not refer to separate networks, but simply to the fact that experience was collected in parallel rollouts. We used the same values as in \citep{NEURIPS2018_e5841df2}, i.e.,
\begin{itemize}
    \item batch size $512$, with $128$ parallel agents, for all datasets. This means that each agent received a batch of size $4$.
\end{itemize}

Furthermore, the total number of batches used was always 10000. Although some methods would have possibly benefited from more training, we wanted to avoid excessive hyperparameter tuning. Instead, by giving the exact same amount of data to each method, sample efficiency became a part of the evaluation.

\subsubsection{JAFA-MFRL}

\textbf{Changes to pretraining/architecture}

\begin{itemize}
    \item Original \method{JAFA} specified the hidden layers for the reading block cells of the set encoder introduced in \citep{vinyalsOrderMattersSequence2016} as $[32,32]$, but not the hidden layers of the writing block. We chose to set this to $[32,32]$ as well.
    \item We used $5$ processing steps for the RNN.
    \item Original \method{JAFA} used a constant masking probability of $0.5$, while ours varied.
    \item Original \method{JAFA} used hidden layers $[32,32]$ for the classifier. We used this for all datasets, except for \dataset{(Fashion)MNIST} where we found that the classifier severely underfitted. In this case, we increased the model complexity to $[128,128]$ instead.
\end{itemize}

\textbf{Changes to RL training}

\begin{itemize}
    \item Original \method{JAFA} used 4-step Q-learning. Since TorchRL does not make this readily available, we instead used a TD($\lambda$) value estimator with $\lambda=0.75$. This value was chosen since it gives the largest weight to the 4-step return (see Appendix \ref{seq:lambda-proof}).
\end{itemize}

Some additional insights gained when training this method:
\begin{itemize}
    \item This is by far the slowest of the RL methods, due to the RNN. See \Cref{appendix:compute}.
    \item Using a (prioritized) replay buffer could improve the method.
    \item We often found that performance (in terms of accuracy) degraded at the start of training. This could be ameliorated by only initiating joint training once $\epsilon$ for the agent has decreased.
\end{itemize}

\subsubsection{ODIN-MBRL}
\textbf{Changes to pretraining/architecture}
\begin{itemize}
    \item Original \method{ODIN} masked features with probability sampled from $\mathcal{U}(0,1)$, we have different probabilities.
    \item The scaling factor for the KL loss term when training the PVAE was set to $0.1$ for all datasets. We tried the default value of $1$ first but found that it led to posterior collapse.
    \item Similarly to \method{JAFA}, hidden layers $[128,128]$ were used for \dataset{(Fashion)MNIST} and $[32,32]$ for the remaining datasets.
\end{itemize}

\textbf{Changes to RL training}
\begin{itemize}
    \item Original \method{ODIN} does not describe the number of parallel agents and the batch size that they use. We used consistent values across all RL methods.
    \item For PPO we used an entropy bonus coefficient of $0.01$, the default value in TorchRL.
\end{itemize}

\subsubsection{ODIN-MFRL}

\method{ODIN-MFRL} was trained in exactly the same way as \method{ODIN-MBRL}, expect that it disabled generative rollouts.

\subsubsection{OL-MFRL+no-mask}

In the original implementation, there was no pretraining stage for training the classifier separately. However, the reward that the agent receives is only as useful as the classifier it is based on, and we found that a majority of the training time was ''wasted'' on waiting for the classifier to improve. Hence, we added a pretraining stage, improving training efficiency.

\textbf{Changes to pretraining/architecture}
\begin{itemize}
    \item We used the same architecture for \dataset{(Fashion)MNIST} as originally presented: $[512,512,128,64]$. For the remaining datasets, we used the architecture originally used for the \dataset{Diabetes} dataset, $[64,32,16]$.
\end{itemize}

\textbf{Changes to RL training}
\begin{itemize}
    \item We did not use an adaptive learning rate.
    \item The \method{OL} repo only provides the number of training episodes for the \dataset{Diabetes} dataset. As for all other RL methods, we pretrained until convergence and trained with a consistent number of batches.
    \item Original \method{OL} updated the target network in discrete steps, but we used a soft update instead, with $\tau = 0.005$.
    \item Original \method{OL} trained a single agent with batch size 1. We tried this, but it made the method very unstable. Consequently, we increased the number of agents and batch size to be consistent across all RL methods.
    \item Original \method{OL} decayed $\epsilon$ exponentially, but we decreased it linearly during the first half of training. This makes it consistent with our \method{JAFA} implementation.
    \item Original \method{OL} prescribed a replay buffer of size $1000 n_\textrm{features}$. We used a fixed size of $100\,000$ instead.
\end{itemize}

Some additional insights gained when training this method:
\begin{itemize}
    \item The architectural simplicity comes at a cost: the method cannot differentiate between unobserved features and observed features with value $0$. This is a significant drawback that makes the method perform especially poorly on datasets with a large number of zeros. Due to this deficiency, we also investigated a variant of \method{OL} which concatenated the feature mask to the input.
\end{itemize}

\subsubsection{OL-MFRL}
\method{OL-MFRL} makes no additional modifications to \method{OL-MFRL+no-mask} apart from concatenating the feature mask to the input. This changes the size of the first layers in both the P-network and Q-network.

\subsubsection{AACO}
Although the \method{AACO} method avoids some problems that arise when using an RL method, to adhere to the fair benchmarking framework, several changes were needed:

\begin{itemize}
    \item The \method{AACO} method cannot execute an entire selected subset in one step; instead, it breaks ties by selecting the feature that minimizes expected loss (without an explicit cost term), as in the original implementation.
    \item In the hard-budget setting, the method cannot terminate early.
    \item To adhere to the fair benchmark, the original classifiers suggested by \citeauthor{pmlr-v235-valancius24a} (i.e., \texttt{XGBClassifier} and Bayes classifier) were not used. Instead, the external classifier was used.
\end{itemize}

\subsubsection{AACO+NN}
For \method{AACO+NN}, we first loaded a trained \method{AACO} object and generated oracle rollouts on the training split. Rollouts used the same \texttt{Initializer} and \texttt{Unmasker} as evaluation. In soft-budget runs, the oracle could stop early; in hard-budget runs, forced acquisition was enabled and the rollout length was capped by the budget.

The policy network was a masked-MLP-style model taking concatenated masked features and feature mask as input, with hidden layers $[256, 256]$, dropout $0.1$, and an output dimension equal to the number of selections plus one stop action. We trained with \texttt{AdamW} (learning rate $10^{-3}$), batch size $256$, 10\% validation split, and early stopping with patience 10 (maximum 100 epochs). The final saved method combines this policy with the same classifier object used for \method{AACO} predictions.

\subsubsection{Common across Discriminative Myopic Methods}
For both discriminative myopic methods, we used the same predictor setup for each modality to ensure a fair comparison. Specifically, for tabular datasets, we used a predictor with two hidden layers of size [128,128] and a dropout rate of 0.3, whereas for Imagenette we used a pretrained ResNet-50 \cite{he2015deepresiduallearningimage} predictor. Both methods involved predictor pretraining and joint training of the predictor and selector/estimator. During training, for tabular datasets, we concatenated the masked features with the corresponding feature mask as the model input according to the default settings of \cite{covert2023learning,gadgil2024estimating}, while for \dataset{Imagenette} we used the masked images directly as the model input. We used a batch size of 128 during both model pretraining and training, and we used early stopping in both cases. 

\subsubsection{GDFS-DM}
We used the default gradual temperature annealing strategy \citep{covert2023learning} to train the models during the joint training phase. For tabular datasets, we used an initial learning rate of $1\times10^{-3}$ and gradually decreased the temperature from 1.0 to 0.1, with a total of five temperature stages using logarithmic spacing. At each temperature stage, we trained the model for a maximum of 250 epochs. We used the model with the lowest validation loss at each temperature as the starting point for the next temperature stage. The checkpoint with the lowest validation loss across all stages was selected as the final model. For \dataset{Imagenette}, we used the same annealing schedule, but with an initial learning rate of $1\times10^{-5}$, a minimum learning rate of $1\times10^{-8}$, and a maximum of 30 epochs per stage.

\subsubsection{DIME-DM} 
During the joint training phase of \method{DIME-DM}, we performed feature selection using the \(\epsilon\)-greedy strategy with the initial \(\epsilon\) set to 0.05. For tabular datasets, if the validation accuracy did not improve over five consecutive epochs, we reduced the \(\epsilon\) by multiplying it by 0.2, up to a maximum of ten times. We also used the default bounded strategy \cite{gadgil2024estimating} for CMI estimation. We jointly trained the model for a maximum of 250 epochs. For \dataset{Imagenette}, we used the same \(\epsilon\)-greedy strategy with the same initial \(\epsilon\) and decay factor, but reduced \(\epsilon\) after three consecutive epochs without validation accuracy improvement and trained the model for up to 150 epochs. For joint training on tabular datasets, we tied the first two linear layers of the predictor and estimator so that they share the same module. For joint training on Imagenette, we used a shared backbone for both the predictor and the estimator.

\subsubsection{EDDI-GM} 
After joint pretraining of the PVAE and predictor, the \method{EDDI-GM} model did not require further training. Hence, we only saved the same models for different budget values and classifier settings in the training stage and used this model for subsequent evaluation. During evaluation, we considered two variants of the \method{EDDI} method in this paper. They shared the same feature selection procedure, but differed in how predictions were computed during feature selection. In the \method{EDDI-GM+builtin} variant, the method performed multiple Monte Carlo samplings based on the current masked features to obtain latent representations, and used the PVAE's built-in classifier to compute the current predictions by averaging over these representations. The same Monte Carlo samples were then used to complete missing features, and candidate inputs were constructed by sequentially revealing each input feature. Predictions were recomputed on these candidate inputs and compared with the original predictions to measure the informativeness of each feature. In contrast, in the \method{EDDI-GM} variant, the original predictions were made directly based on the masked features, and the sampler was only used to complete missing feature values. During feature informativeness evaluation, predictions were computed directly on the completed candidate inputs without resampling latent representations. We used a total of 128 Monte Carlo samples in both variants.

\subsubsection{Common across Static Methods}
During the training phase of the static methods on tabular datasets, we used a selector and classifier with two hidden layers of size [128,128] for both methods. We trained both methods with a batch size of 128 and a learning rate of 0.001.

\subsubsection{CAE-S}
For tabular datasets experiments, we trained the selector for up to 2500 epochs and the predictor for up to 250 epochs, both with an early stopping strategy. For \dataset{Imagenette} experiments, we used a pretrained ResNet-50 \cite{he2015deepresiduallearningimage} for both the selector and classifier. The selector was trained for up to 500 epochs and the classifier was trained for up to 50 epochs. Both the selector and the classifier were trained with a learning rate of $1\times10^{-5}$. For the selector training on both tabular and image datasets, we used 10 Gumbel-Softmax temperature annealing steps.

\subsubsection{PT-S}
For tabular dataset experiments, we trained both the selector and the classifier for a maximum of 250 epochs with an early stopping strategy.

\subsection{Hardware}

\begin{itemize}
    \item Myopic methods on the \dataset{Imagenette} dataset used NVIDIA A40 GPUs during the training stage.
    \item AACO(+NN) used 4 AMD EPYC 9354 Zen4 CPU cores during both training and evaluation.
    \item The remaining methods used one NVIDIA T4 GPU during both (pre)training and evaluation.
\end{itemize}


\subsection{Proof that $\lambda=0.75$ gives maximum weight to 4-step return} \label{seq:lambda-proof}
The $\lambda$-return that TD($\lambda$) \citep{Sutton1988} estimates is
\begin{equation}
    G_t^\lambda \doteq (1-\lambda)\sum_{n=1}^\infty \lambda^{n-1}G_{t:t+n}
\end{equation}
where $G_{t:t+n}$ is the n-step return,
\begin{equation}
    G_{t:t+n} \doteq R_{t+1} + \gamma R_{t+2} + \dots + \gamma^{n-1} R_{t+n} + \gamma^n Q(S_{t+n}, A_{t+n}) .
\end{equation}

The weight given to the 4-step return $G_{t:t+4}$ in $G_t^\lambda$ is
\begin{equation}
    w_4 = (1-\lambda)\lambda^3
\end{equation}
which when differentiated,
\begin{equation}
    \frac{\partial}{\partial \lambda}w_4 = \lambda^2(3-4\lambda)
\end{equation}
shows that $w_4$ is maximized by $\lambda=0.75$. $\qquad\qquad\qquad\qquad\square$

\section{Proof of Theorem \ref{thm:cube-nm}} \label{appendix:proofcube}

We begin by stating the theorem more formally. Note that we also include the worst-case and best-case cost here.

\begin{theorem}[Noiseless CUBE-NM query complexity]\label{thm:cube-nm-formal}
Consider the CUBE-NM dataset with $n_c$ contexts in the noiseless setting $\sigma=0$, so that each instance $x\in\mathcal{X}$ determines its label with full certainty (i.e., $H(\mathbf{y}\mid x)=0$). Let $f(S,x_S)=p(\mathbf{y}\mid x_S)$ be Bayes optimal and assume uniform feature costs $c_i=1$ for all $i\in[d]$. In addition, assume we allow any policy to acquire features until $H(\mathbf{y}\mid x_S)=0$, and denote $\Pi$ as the set of all such policies. Let $\pi^\ast = \argmin_{\pi \in \Pi}\mathbb{E}_{\mathbf{x}}\mathbb{E}_{\pi}[c(\pi[\mathbf{x}])]$ denote an optimal policy in terms of expected cost and let $\pi_{\mathrm{CMI}} \in \Pi$ denote the myopic CMI policy in \eqref{eq:cmi}. Then $\mathbb{E}_{\mathbf{x}}\mathbb{E}_{\pi^\ast}[c(\pi^\ast[\mathbf{x}])] = \mathbf{1}\{n_c\ge 2\}+2.25$, with $\min_{x\in\mathcal{X}} c(\pi^\ast[x])=\mathbf{1}\{n_c\ge 2\}+1$ and $\max_{x\in\mathcal{X}} c(\pi^\ast[x])=\mathbf{1}\{n_c\ge 2\}+3$. In contrast, $\mathbb{E}_{\mathbf{x}}\mathbb{E}_{\pi_{\mathrm{CMI}}}[c(\pi_{\mathrm{CMI}}[\mathbf{x}])] = 13(2n_c+1)/16$, with $\min_{x\in\mathcal{X}} c(\pi_{\mathrm{CMI}}[x])=1$ and $\max_{x\in\mathcal{X}} c(\pi_{\mathrm{CMI}}[x])=3n_c$.
\end{theorem}

\begin{proof}
Let $n=n_c$. In the noiseless setting $\sigma=0$, each label-context pair corresponds to a deterministic feature vector.

\textbf{Notation.}
For noiseless \dataset{CUBE}, let $x^{(y)}\in\{0,0.5,1\}^{10}$ denote the 10-dimensional prototype for class $y\in\{1,\dots,8\}$ (restricted to the first 10 coordinates), and let $x^{(y)}_i$ be its $i$-th coordinate.
For \dataset{CUBE-NM} with $n$ contexts, each instance has one context feature $f_1$ with value $x_{f_1}\in\{1,\dots,n\}$, and $n$ disjoint 10-dimensional blocks. Write $g_{b,i}$ for the $i$-th feature inside block $b\in\{1,\dots,n\}$. If the label is $y$ and the active block is $j$, then
$x_{g_{j,i}}=x^{(y)}_i$ for all $i\in\{1,\dots,10\}$ and $x_{g_{b,i}}=0.5$ for all $b\neq j$.
Unit costs mean $c(\pi[x])$ equals the number of queried features.

\textbf{Step 1: noiseless \dataset{CUBE} costs.}
Consider the noiseless \dataset{CUBE} core on 10 features, with 8 classes and deterministic prototypes $x^{(y)}\in\{0,0.5,1\}^{10}$ for $y\in\{1,\dots,8\}$. A policy adaptively queries feature indices $i\in\{1,\dots,10\}$ and observes $x^{(y)}_i$.

\emph{An upper bound (a policy with expected cost $2.25$).}
Query feature $7$.
If $x^{(y)}_7=1$, then $y=5$ and stop.
If $x^{(y)}_7=0$, then $y\in\{6,7\}$ and querying feature $6$ separates them.
If $x^{(y)}_7=0.5$, then $y\in\{1,2,3,4,8\}$; query feature $2$, which isolates $y=1$ and $y=2$, and leaves $y\in\{3,4,8\}$, which is resolved by one additional query.
Under the uniform class prior, the expected number of queries is
$1+(2/8)\cdot 1+(5/8)\cdot(1+(3/5)\cdot 1)=2.25$.
The best case is $1$ query (when $x^{(y)}_7=1$), and the worst case is $3$ queries.

\emph{Optimality (dynamic programming on belief states).}
A noiseless state is fully described by a pair $(B,S)$, where $B\subseteq\{1,\dots,8\}$ is the set of labels still consistent with the observed feature values, and $S\subseteq\{1,\dots,10\}$ is the set of already queried features. Let $V(B,S)$ denote the minimum expected number of \emph{additional} queries needed to identify the label when the queried set is $S$ and remaining labels are $B$ (note that the outcomes of the features are irrelevant, because $(B,S)$ is a sufficient statistic at any state for this simple dataset, see \citep{aronsson2025surveyactivefeatureacquisition} for more on this). The boundary condition is $V(\{y\},S)=0$ for all $y$ and $S$.

If a feature $i\notin S$ is queried next and the observed value is $v\in\{0,0.5,1\}$, the remaining possible labels become
$B_{i=v}=\{y\in B: x^{(y)}_i=v\}$ and the queried set becomes $S\cup\{i\}$.
In the noiseless uniform setting, $\Pr(v\mid B,S,i)=|B_{i=v}|/|B|$.
Therefore, for $|B|\ge 2$,
\[
V(B,S)
=1+\min_{i\in\{1,\dots,10\}\setminus S}
\sum_{v\in\{0,0.5,1\}}
\Pr(v\mid B,S,i)\,V(B_{i=v},S\cup\{i\}).
\]
Evaluating this recursion on the noiseless \dataset{CUBE} dataset yields
$V(\{1,\dots,8\},\emptyset)=2.25$ (note that this is straightforward due to the simplicity of the dataset and $\sigma=0$).
Since the policy above attains expected cost $2.25$, it is optimal.
Moreover, the same recursion implies that the best-case cost is $1$ and the worst-case cost is $3$.

\textbf{Step 2: optimal policy on noiseless \dataset{CUBE-NM}.}
If $n=1$, \dataset{CUBE-NM} reduces to the noiseless \dataset{CUBE} dataset (because dummy features are irrelevant with $\sigma=0$), so $\mathbb{E}_{\mathbf{x}}\mathbb{E}_{\pi^\ast}[c(\pi^\ast[\mathbf{x}])]=2.25$, $\min_{x\in\mathcal{X}}c(\pi^\ast[x])=1$, and $\max_{x\in\mathcal{X}}c(\pi^\ast[x])=3$.

Assume $n\ge 2$.
Querying $f_1$ reveals the active block in exactly one query. After that, the remaining task is exactly noiseless \dataset{CUBE} on the active block, which has optimal expected cost $2.25$ by Step 1. Hence the policy ``query $f_1$ first, then run an optimal noiseless \dataset{CUBE} policy inside the active block'' has expected cost $1+2.25$, best case $1+1$, and worst case $1+3$.

This is optimal: any query in an inactive block always returns $0.5$ and provides no information about $y$, so such queries can only increase cost. Since $n\ge 2$, any policy must determine which block is active before it can reliably use informative features, and querying $f_1$ does this in one step. Therefore
$\mathbb{E}_{\mathbf{x}}\mathbb{E}_{\pi^\ast}[c(\pi^\ast[\mathbf{x}])]=\mathbf{1}\{n\ge 2\}+2.25$,
$\min_{x\in\mathcal{X}}c(\pi^\ast[x])=\mathbf{1}\{n\ge 2\}+1$,
and $\max_{x\in\mathcal{X}}c(\pi^\ast[x])=\mathbf{1}\{n\ge 2\}+3$.

\textbf{Step 3: myopic CMI policy.}
Fix $n=n_c$. Recall that $\dataset{CUBE\text{-}NM}$ has one context feature $f_1$ (encoding the active context) and $n$ blocks of 10 features. Denote the block features by $g_{b,i}$ for block $b\in\{1,\dots,n\}$ and within-block index $i\in\{1,\dots,10\}$. In the noiseless setting, if block $b$ is inactive then $\mathbf{x}_{g_{b,i}}=0.5$ deterministically, while in the active block $j$ we observe $\mathbf{x}_{g_{j,i}}=x^{(y)}_i$.

\emph{Intuition for the behavior of $\pi_{\mathrm{CMI}}$.}
By construction the context is independent of the label, so $I(\mathbf{y};\mathbf{x}_{f_1})=0$ and $\pi_{\mathrm{CMI}}$ does not query $f_1$ as long as there exist block features with strictly positive CMI. Initially, the CMI maximizers are the four within-block indices $i\in\{4,5,6,7\}$ in any block.
After querying some $g_{b,i}$ and observing $0.5$, block $b$ becomes less likely to be the active block than any untouched block, because the outcome $0.5$ is guaranteed if $b$ is inactive but occurs only for a strict subset of labels if $b$ is active. Consequently, it can be verified (by a direct CMI calculation) that $\pi_{\mathrm{CMI}}$ prefers to query the same within-block index $i$ in an untouched block rather than querying a different feature in the same block. In effect, at a fixed within-block index $i$, $\pi_{\mathrm{CMI}}$ \emph{tests blocks} one by one until it either observes a non-$0.5$ value (which can only happen in the active block) or it has tested all $n$ blocks and concludes that the active block also has value $0.5$ at index $i$. Once the value at index $i$ in the active block is determined, $\pi_{\mathrm{CMI}}$ moves to the next within-block index that maximizes CMI under the reduced label set, and repeats the same block-testing behavior. The exact comparisons of CMI are straightforward to verify for the noiseless prototypes, so they are omitted for brevity.

\emph{Expected cost.}
The policy $\pi_{\mathrm{CMI}}$ selects, at each step, a feature with maximal CMI
$I(\mathbf{y};\mathbf{x}_a\mid \mathbf{x}_S)$, breaking ties uniformly at random. As argued above, its behavior can be viewed as a sequence of \emph{stages}. In a stage, $\pi_{\mathrm{CMI}}$ fixes a within-block index $i$ (chosen as a CMI maximizer under the current posterior on $\mathbf{y}$) and then queries features of the form $g_{b,i}$ across blocks $b$ until it has determined the value of the active block at index $i$ (either by observing a non-$0.5$ value in some block, or by exhausting all $n$ blocks and concluding that the active block also has value $0.5$). After this, the posterior on $\mathbf{y}$ is reduced, a new maximizer $i$ is selected, and the next stage begins.

\paragraph{Block-testing cost within a stage.}
Fix a stage index $i$ and let
\[
p \;:=\; \Pr\!\big(x^{(\mathbf{y})}_i \neq 0.5 \,\big|\, \text{current posterior on }\mathbf{y}\big).
\]
With probability $p$, the active block produces a non-$0.5$ value at index $i$, and $\pi_{\mathrm{CMI}}$ stops the stage as soon as it tests the active block. With uniform random tie-breaking across blocks, the position of the active block among the tested blocks is uniform on $\{1,\dots,n\}$, so the expected number of block tests in this case is $(n+1)/2$. With probability $1-p$, the active block also has value $0.5$ at index $i$, so every queried block returns $0.5$ and $\pi_{\mathrm{CMI}}$ must test all $n$ blocks to conclude this. Therefore, the expected number of queries spent in this stage is
\[
q(p) \;:=\; p\cdot\frac{n+1}{2} \;+\; (1-p)\cdot n.
\]

\paragraph{First-stage tie and two symmetry classes.}
Initially, under the uniform prior on $\mathbf{y}\in\{1,\dots,8\}$, the CMI maximizers are exactly the indices $i\in\{4,5,6,7\}$, and the first queried feature is uniformly random among $\{g_{b,i}: b\in\{1,\dots,n\},\, i\in\{4,5,6,7\}\}$. By symmetry, the event $i\in\{4,7\}$ occurs with probability $1/2$ and the event $i\in\{5,6\}$ occurs with probability $1/2$. These two cases lead to different reduced label sets after the first stage, and hence to different expected remaining costs. Let $G(n)$ denote the expected total cost conditional on the first-stage index belonging to $\{4,7\}$, and let $B(n)$ denote the analogous quantity conditional on the first-stage index belonging to $\{5,6\}$.

\paragraph{Case A: first stage uses $i\in\{4,7\}$.}
By symmetry it suffices to analyze $i=4$. From the noiseless $\dataset{CUBE}$ prototypes, exactly $3$ of the $8$ labels satisfy $x^{(y)}_4\neq 0.5$, so the first stage has $p=3/8$ and expected cost $q(3/8)$. If the active-block value is non-$0.5$, then with probability $2/3$ of that event the observation corresponds to two labels, which requires one additional query to resolve; this contributes $(2/8)\cdot 1$ to the expectation. If instead the active-block value is $0.5$ (probability $5/8$), the posterior reduces to a 5-label set, and the next myopic maximizer is $i=7$. On this 5-label set, exactly $3$ labels satisfy $x^{(y)}_7\neq 0.5$, so the second stage has $p=3/5$ and expected block-testing cost $q(3/5)$. If this stage yields a non-$0.5$ value in the active block, then with probability $2/3$ of that event two labels remain and one further query resolves the label; this contributes $(2/5)\cdot 1$ within the 5-label branch. If instead the active-block value is again $0.5$ (probability $2/5$ within the 5-label branch), then exactly two labels remain and the final distinguishing stage has $p=1/2$, contributing $q(1/2)$. Altogether,
\[
\begin{aligned}
G(n)
&= q(3/8) + (2/8)\cdot 1 + (5/8)\,H(n),\\
H(n)
&= q(3/5) + (2/5)\cdot 1 + (2/5)\,q(1/2),
\end{aligned}
\]
which simplifies to $G(n)=(23n+15)/16$.

\paragraph{Case B: first stage uses $i\in\{5,6\}$.}
By symmetry it suffices to analyze $i=5$. Again exactly $3$ of the $8$ labels satisfy $x^{(y)}_5\neq 0.5$, so the first stage has $p=3/8$ and cost $q(3/8)$, and the non-$0.5$ outcome yields a two-label ambiguity with probability $2/8$, contributing $(2/8)\cdot 1$. If the active-block value is $0.5$ (probability $5/8$), the posterior reduces to a 5-label set and the next myopic maximizer is $i=2$, for which exactly $2$ of the $5$ labels satisfy $x^{(y)}_2\neq 0.5$; hence the next stage has $p=2/5$ and cost $q(2/5)$. If the active-block value at $i=2$ is non-$0.5$, then one additional query resolves the remaining two labels; if it is $0.5$ (probability $3/5$ within this branch), then the posterior reduces to a 3-label set. On that 3-label set, the myopic continuation performs a stage with $p=1/3$ (detecting a single label) and, if needed, a final stage with $p=1/2$ (separating the last two labels). Thus
\[
\begin{aligned}
B(n)
&= q(3/8) + (2/8)\cdot 1 + (5/8)\,M(n),\\
M(n)
&= q(2/5) + (3/5)\,C(n),\\
C(n)
&= (1/3)\cdot\frac{n+1}{2} + (2/3)\Big(n + q(1/2)\Big)
= \frac{4n+1}{3},
\end{aligned}
\]
which simplifies to $B(n)=(29n+11)/16$.

\paragraph{Averaging over the initial tie.}
Finally, averaging over the first-stage symmetry classes (each with probability $1/2$) gives
\[
\mathbb{E}_{\mathbf{x}}\mathbb{E}_{\pi_{\mathrm{CMI}}}\!\big[c(\pi_{\mathrm{CMI}}[\mathbf{x}])\big]
=\frac{G(n)+B(n)}{2}
=\frac{13(2n+1)}{16}.
\]

\emph{Best-case cost.}
There exist labels $y$ for which a single informative query in the active block uniquely identifies the label (for example, at index $i=4$ the value $0$ occurs for exactly one class). Since $\pi_{\mathrm{CMI}}$ may select such a feature in the active block on the first step, $\min_{x\in\mathcal{X}} c(\pi_{\mathrm{CMI}}[x])=1$.

\emph{Worst-case cost.}
There exist labels $y$ for which three successive myopic stages require exhausting all $n$ blocks before a non-$0.5$ value is observed or the stage is concluded (e.g., the active block yields $0.5$ at each of the three within-block indices selected by the myopic rule). In that event each stage costs $n$ block tests, so $\max_{x\in\mathcal{X}} c(\pi_{\mathrm{CMI}}[x])=3n$.
\end{proof}

\end{document}